%% file: manuscript.tex
\documentclass[11pt]{article}

\usepackage[margin=1in]{geometry}
\usepackage[T1]{fontenc}
\usepackage[utf8]{inputenc}
\usepackage{lmodern}
\usepackage{microtype}
\usepackage{array}
\usepackage{booktabs}
\usepackage{enumitem}
\usepackage{graphicx}
\usepackage{xcolor}
\usepackage{tabularx}
\usepackage{placeins}
\usepackage{tikz}
\usepackage[most]{tcolorbox}
\usepackage{subcaption}
\usepackage[numbers,sort&compress]{natbib}
\usepackage[colorlinks=true,linkcolor=blue,citecolor=blue,urlcolor=blue]{hyperref}
\usepackage{makecell}
\newcolumntype{T}{>{\hsize=3.20\hsize\linewidth=\hsize\raggedright\arraybackslash}X}
\newcolumntype{U}{>{\hsize=0.70\hsize\linewidth=\hsize\centering\arraybackslash}X}
\newcolumntype{V}{>{\hsize=0.90\hsize\linewidth=\hsize\centering\arraybackslash}X}

\usetikzlibrary{arrows.meta,calc,positioning,decorations.pathreplacing}

\usepackage{fontawesome5}

\newtcolorbox{evidencebox}[1]{
  enhanced,
  breakable,
  colback=gray!2,
  colframe=black!18,
  colbacktitle=gray!10,
  coltitle=black,
  fonttitle=\bfseries\small,
  title={#1},
  boxrule=0.45pt,
  arc=2pt,
  left=6pt,
  right=6pt,
  top=5pt,
  bottom=5pt,
  before skip=6pt,
  after skip=6pt
}

\definecolor{taBlue}{HTML}{2E78B7}
\definecolor{taCyan}{HTML}{3E9BC6}
\definecolor{taRed}{HTML}{D65F4C}
\definecolor{taSky}{HTML}{77B7D5}
\definecolor{TopicRed}{RGB}{160, 82, 68}
\definecolor{MainGray}{RGB}{120, 120, 120}
\definecolor{FlowGray}{RGB}{105, 105, 105}
\definecolor{ArchBlue}{RGB}{34, 103, 156}
\definecolor{ArchFill}{RGB}{245, 250, 253}
\definecolor{AcqTeal}{RGB}{43, 135, 144}
\definecolor{AcqFill}{RGB}{244, 251, 250}

\newcolumntype{Y}{>{\raggedright\arraybackslash}X}
\newcolumntype{C}{>{\centering\arraybackslash}p{1.05cm}}

\newenvironment{promptbox}[1]{%
  \begin{evidencebox}{#1}
}{%
  \end{evidencebox}
}

\newcommand{\compactentry}[1]{%
  \par\smallskip\noindent\textbf{#1}\enspace
}

\newenvironment{caseentry}[1]{%
  \par\vspace{4pt}\noindent
  \begin{minipage}{\columnwidth}
  \textbf{#1}\par\vspace{2pt}
}{%
  \end{minipage}\par\vspace{4pt}
}

\title{\bfseries Terminal Agents: A Survey of AI Agents in Command-Line Environments}

\author{%
  Yi Bin\textsuperscript{1}\thanks{These authors contributed equally to this research.}\quad
  Xiaoyang Yuan\textsuperscript{1,4}\footnotemark[1]\quad
  Haoxi Zeng\textsuperscript{1}\footnotemark[1]\quad
  Wencheng Ye\textsuperscript{1}\footnotemark[1]\\[0.35em]
  Wenqi Shao\textsuperscript{4}\quad
  Chen Qian\textsuperscript{2}\quad
  Wei Ye\textsuperscript{1}\quad
  Yujuan Ding\textsuperscript{3}\\[0.35em]
  Zheng Wang\textsuperscript{1}\quad
  Pengpeng Zeng\textsuperscript{1}\quad
  Jingkuan Song\textsuperscript{1}\quad
  Heng Tao Shen\textsuperscript{1}\\[0.45em]
  \normalsize \textsuperscript{1}Tongji University, Shanghai, China\\
  \normalsize \textsuperscript{2}Shanghai Jiao Tong University, Shanghai, China\\
  \normalsize \textsuperscript{3}The Hong Kong Polytechnic University, Hong Kong, China\\
  \normalsize \textsuperscript{4}Shanghai Innovation Institute, Shanghai, China
}

\date{}

\begin{document}

\maketitle

\begin{abstract}
\input{sections/0_abstract}
\end{abstract}

\vspace{-0.4em}
\begin{center}
\small
\faIcon{github} \textbf{GitHub:}
\href{https://github.com/EnigmaYYYY/awesome-terminal-agents}
{\texttt{https://github.com/EnigmaYYYY/awesome-terminal-agents}}
\end{center}
\vspace{-0.2em}

\input{sections/1_introduction}
\input{sections/2_background_terminal_agents}
\input{sections/3_technical_evolution_system_architectures}
\input{sections/4_terminal_competence_acquisition}
\input{sections/5_benchmarks_metrics_evaluation}
\input{sections/6_empirical_analysis_insights}
\input{sections/7_challenges_future}
\input{sections/8_conclusion}

\bibliographystyle{plainnat}
\bibliography{references}

\appendix
\section*{APPENDIX}
\input{sections/appendix_methodology}

\end{document}

%% file: sections/0_abstract.tex
Large language model agents increasingly act through terminals, yet existing surveys disperse terminal-mediated behavior across software engineering, tool use, and computer-use research. We regard terminal agents as systems whose dominant progress-bearing action--observation loop is mediated by terminal command execution, textual feedback, and stateful environment interaction. Using terminal-mediated execution as an organizing lens, this survey establishes workload-level boundaries and connects system architecture, competence acquisition, and evaluation through a seven-dimensional terminal competence profile. Our synthesis shows that realized behavior is jointly shaped by the model, interface, harness, runtime, and environment. Executable trajectories ground learning in action consequences, verification, and recovery, whereas prevailing evaluations emphasize final outcomes and expose process quality, recovery, and governance unevenly. Bounded fixed-condition diagnostics illustrate two implications: benchmark families expose different process signals, and matched system comparisons reveal benchmark-dependent performance and limits of component attribution. These findings motivate explicit reporting of system and runtime conditions, supported by replayable traces and process-level evidence. The framework provides a unified basis for studying terminal-mediated agency across software engineering and emerging application domains.

%% file: sections/1_introduction.tex
\section{Introduction}
\label{sec:introduction}

Large language models (LLMs) are evolving from prompt-bound text generators into interactive systems that act in external environments and revise their behavior from feedback~\cite{yao2022react,schick2023toolformer,wang2024executable}. Through iterative action and observation, these systems increasingly operate as agents that execute code, invoke computational tools, navigate graphical interfaces, and modify digital environments over multiple steps~\cite{parisi2022talm,hu2025agents}. The medium through which such agents act shapes their available actions, observable state, feedback, and means of verification, and is therefore part of the agent system rather than a neutral implementation channel.

\begin{figure*}[!t]
\centering
\includegraphics[width=0.8\textwidth]{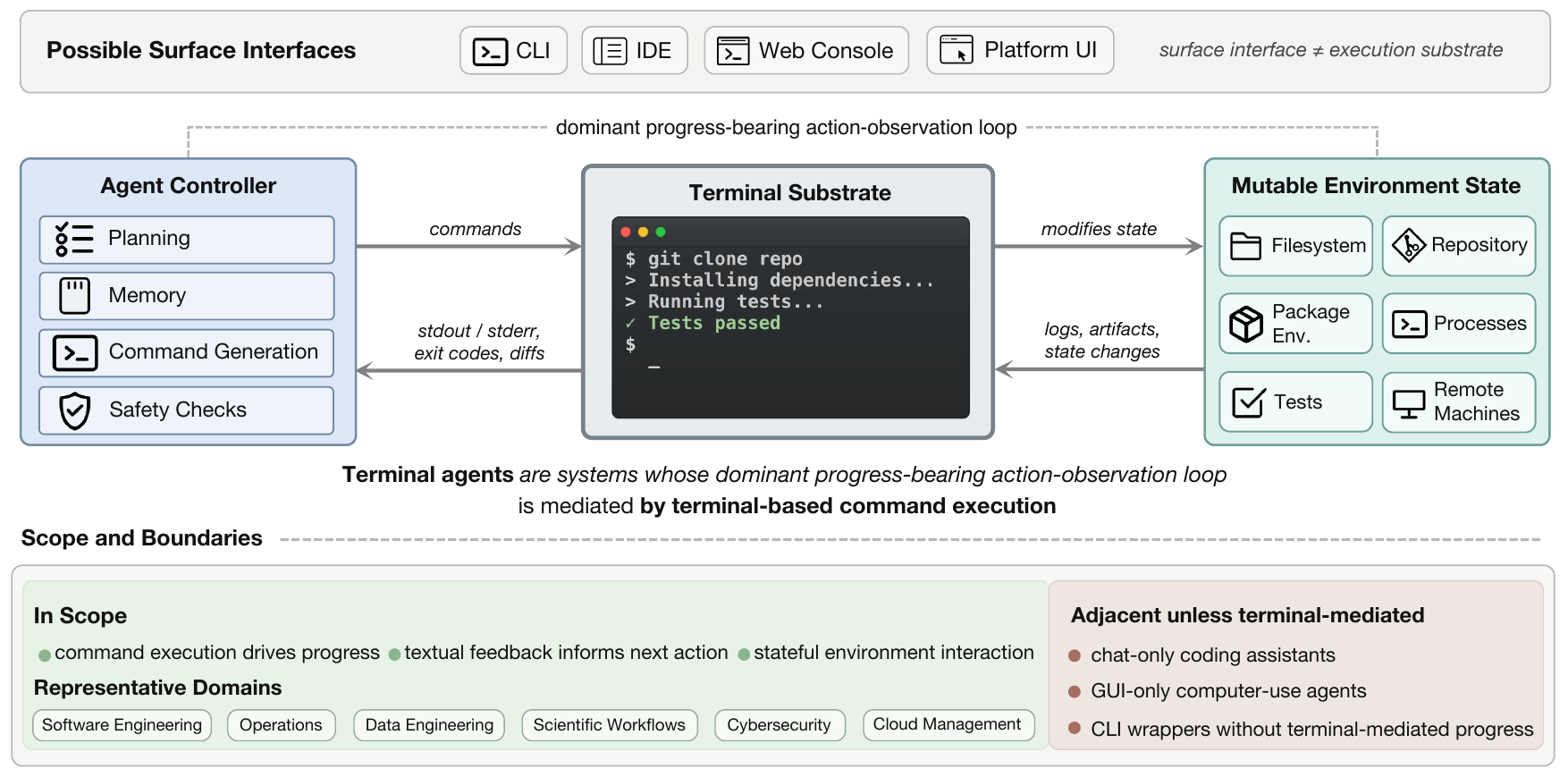}
\caption{Terminal agents and related concepts. An agent controller, terminal substrate, and mutable environment state form a progress-bearing interaction loop. Systems are in scope when command execution drives progress, textual feedback guides later actions, and stateful environment interaction is central; surface-level CLI access and incidental command use remain adjacent.}
\label{fig:terminal_agent_definition}
\end{figure*}

The terminal is a particularly consequential execution medium because it provides compact, scriptable textual access to stateful runtimes comprising filesystems, dependencies, processes, tests, logs, remote machines, and command-line tools. Commands can modify files, configure environments, launch services, and execute tests, while outputs, exit codes, diffs, stack traces, logs, and artifacts inform subsequent decisions. Terminal-mediated execution thus couples reasoning, execution, observation, and verification through a mutable action--observation loop~\cite{de2026terminal}. When this loop is the principal mechanism of task progress, the terminal becomes an execution substrate rather than merely an access interface. We accordingly regard systems whose dominant progress-bearing action--observation loop is mediated by terminal command execution, textual feedback, and stateful environment interaction as \emph{terminal agents}. Figure~\ref{fig:terminal_agent_definition} distinguishes this substrate from surface interfaces such as CLIs, IDEs, and web consoles.

This loop makes several general capabilities jointly inspectable in execution traces: planning appears as executable action sequences, memory is tested against persistent state, adaptation is grounded in external feedback, and verification occurs in the same substrate. Terminal agents therefore provide a concrete setting for studying interactive, environment-grounded intelligence. Existing surveys, however, organize related evidence around general LLM agents~\cite{wang2024survey,xi2025rise}, software-engineering agents~\cite{wang2025agents,wang2025ai}, GUI- and browser-based computer use~\cite{hu2025agents,shi2025towards}, vision-language-action models~\cite{yu2025survey}, or agent evaluation~\cite{yehudai2025survey,dong2025survey}. Terminal-mediated behavior consequently remains dispersed across autonomy, repository repair, computer use, and evaluation, without a common synthesis of scope, system responsibilities, acquisition pathways, and measurement problems.

The literature supporting such a synthesis remains uneven across domains. Software engineering provides the densest empirical foundation because repositories, tests, and build systems make terminal-mediated behavior directly observable through executable outcomes~\cite{jimenez2024swe,yang2024swe,merrill2026terminal}. Benchmarks, executable training environments, post-training studies, and deployment reports increasingly treat terminal interaction as an explicit design and evaluation target rather than a passive backend~\cite{merrill2026terminal,lin2026cli,pan2024training,bechard2026terminal}. Related terminal-mediated interaction also appears in operations, data engineering, scientific workflows, cybersecurity, and cloud management, but evidence remains fragmented~\cite{jha2025itbench,chen2025aiopslab,jin2025elt,kon2025curie,liu2025pacebench,parthasarathy2025engineering}. We therefore distinguish established findings from emerging practices; the Supplementary Material reports the review protocol, corpus construction, and evidence-calibration procedure.

Against this background, terminal-mediated execution provides the organizing lens, while the seven-dimensional terminal competence profile supplies a common analytical language. The survey makes three connected contributions:
\begin{itemize}[leftmargin=*,itemsep=1pt,topsep=2pt]

\item \textbf{Operational scope and boundaries.} We provide a substrate-centered characterization and workload-level tests separating progress-bearing terminal execution from surface-level CLI access, incidental command use, and adjacent interaction substrates.

\item \textbf{Integrated analytical framework.} We connect system architecture, competence acquisition, and evaluation through a seven-dimensional terminal competence profile for comparing system responsibilities, learning signals, and observable evidence.

\item \textbf{Cross-cutting synthesis and diagnostics.} Applying this framework, we synthesize how the model, interface, harness, runtime, and environment jointly shape terminal-agent behavior, how executable trajectories support competence acquisition, and how prevailing evaluations expose process behavior unevenly. Bounded fixed-condition diagnostics further illustrate benchmark-dependent process exposure and limits of component attribution.

\end{itemize}

\input{figures/article_organization}

Figure~\ref{fig:article_organization} summarizes this progression. Section~\ref{sec:background_terminal_agents} establishes the operational scope and terminal competence profile. Sections~\ref{sec:terminal_agent_system_design_architectures} through~\ref{sec:benchmarks_metrics_evaluation} examine how terminal competence is distributed across system architecture, acquired from executable interaction, and observed through evaluation protocols. Section~\ref{sec:cross_cutting_insights} illustrates benchmark-dependent process exposure and attribution limits through bounded fixed-condition diagnostics. Section~\ref{sec:challenges_future} presents the research agenda, and Section~\ref{sec:conclusion} concludes.

%% file: figures/article_organization.tex

\definecolor{TAred}{RGB}{165,84,67}
\definecolor{TAgray}{RGB}{112,116,122}

\definecolor{TAblue}{RGB}{40,110,170}
\definecolor{TAbluefill}{RGB}{247,250,253}

\definecolor{TAteal}{RGB}{45,145,155}
\definecolor{TAtealfill}{RGB}{247,251,251}

\definecolor{TAorange}{RGB}{181,126,56}
\definecolor{TAorangefill}{RGB}{253,250,244}

\definecolor{TAredfill}{RGB}{252,248,247}

\definecolor{TAumbrella}{RGB}{78,82,87}
\definecolor{TAumbrellafill}{RGB}{249,249,248}

\begin{figure*}[!t]
\centering

\resizebox{0.99\textwidth}{!}{%
\begin{tikzpicture}[
    x=1cm,
    y=1cm,
    >=Latex,
    font=\small,
    every node/.append style={
        execute at begin node={
            \hyphenpenalty=10000
            \exhyphenpenalty=10000
        }
    },
%
    basebox/.style={
        rectangle,
        rounded corners=1.2pt,
        align=center,
        line width=0.58pt,
        inner xsep=3.5pt,
        inner ysep=2.7pt,
        outer sep=0pt,
        fill=white
    },
    topicbox/.style={
        basebox,
        draw=TAumbrella,
        fill=TAumbrellafill,
        text width=5.00cm,
        minimum height=0.58cm,
        line width=0.72pt,
        font=\bfseries\fontsize{7.8}{8.8}\selectfont
    },
%
    stagebase/.style={
        basebox,
        text width=2.88cm,
        minimum height=1.16cm,
        font=\fontsize{6.55}{7.40}\selectfont
    },
    stageFound/.style={
        stagebase,
        draw=TAgray,
        fill=black!0.8
    },
    stageArch/.style={
        stagebase,
        draw=TAblue,
        fill=TAbluefill
    },
    stageAcq/.style={
        stagebase,
        draw=TAteal,
        fill=TAtealfill
    },
    stageEval/.style={
        stagebase,
        draw=TAorange,
        fill=TAorangefill
    },
    stageAgenda/.style={
        stagebase,
        draw=TAred,
        fill=TAredfill
    },
%
    itemBase/.style={
        basebox,
        text width=2.72cm,
        minimum height=0.54cm,
        font=\fontsize{5.90}{6.75}\selectfont
    },
    itemFound/.style={
        itemBase,
        draw=TAgray!78,
        fill=black!0.5
    },
    itemArch/.style={
        itemBase,
        draw=TAblue!84,
        fill=TAbluefill
    },
    itemAcq/.style={
        itemBase,
        draw=TAteal!84,
        fill=TAtealfill
    },
    itemEval/.style={
        itemBase,
        draw=TAorange!84,
        fill=TAorangefill
    },
    itemAgenda/.style={
        itemBase,
        draw=TAred!76,
        fill=TAredfill
    },
%
    referenceCell/.style={
        rectangle,
        align=center,
        text width=2.82cm,
        minimum height=0.88cm,
        inner xsep=1.8pt,
        inner ysep=1.4pt,
        outer sep=0pt,
        line width=0pt,
        font=\fontsize{4.60}{5.15}\selectfont,
        text=black!78
    },
%
    umbrellaline/.style={
        draw=TAumbrella!82,
        line width=0.68pt,
        line cap=round,
        line join=round
    },
%
    flowarrow/.style={
        -{Latex[length=1.15mm,width=0.90mm]},
        draw=black!46,
        line width=0.54pt,
        shorten <=1.0pt,
        shorten >=1.0pt
    },
%
    foundline/.style={
        draw=TAgray!80,
        line width=0.62pt
    },
    archline/.style={
        draw=TAblue,
        line width=0.64pt
    },
    acqline/.style={
        draw=TAteal,
        line width=0.64pt
    },
    evalline/.style={
        draw=TAorange,
        line width=0.64pt
    },
    agendaline/.style={
        draw=TAred,
        line width=0.64pt
    }
]


\def\xFound{0.00}
\def\xArch{3.42}
\def\xAcq{6.84}
\def\xEval{10.26}
\def\xAgenda{13.68}

\def\yUmbrella{0.80}

\def\yOne{-1.46}
\def\yTwo{-2.20}
\def\yThree{-2.94}
\def\yFour{-3.68}

\def\yRoute{-0.90}

\def\yRefTitle{-4.31}
\def\yReference{-5.00}


\node[topicbox] (topic) at (\xAcq,1.38)
    {Terminal Agents};


\node[stageFound] (foundation) at (\xFound,0)
    {\textbf{Operational Foundation}\\
     {\color{black!72}
      \fontsize{5.18}{5.85}\selectfont
      Scope, boundaries, and\\
      competence profile}\\
     {\fontsize{5.05}{5.75}\selectfont
      \textbf{Section II}}};

\node[stageArch] (architecture) at (\xArch,0)
    {\textbf{System Architecture}\\
     {\color{black!72}
      \fontsize{5.18}{5.85}\selectfont
      Allocation of\\
      system responsibilities}\\
     {\fontsize{5.05}{5.75}\selectfont
      \textbf{Section III}}};

\node[stageAcq] (acquisition) at (\xAcq,0)
    {\textbf{Competence Acquisition}\\
     {\color{black!72}
      \fontsize{5.18}{5.85}\selectfont
      Learning and adaptation\\
      from executable trajectories}\\
     {\fontsize{5.05}{5.75}\selectfont
      \textbf{Section IV}}};

\node[stageEval] (evaluation) at (\xEval,0)
    {\textbf{Evaluation and Diagnostics}\\
     {\color{black!72}
      \fontsize{5.18}{5.85}\selectfont
      Process observability\\
      and attribution limits}\\
     {\fontsize{5.05}{5.75}\selectfont
      \textbf{Sections V--VI}}};

\node[stageAgenda] (agenda) at (\xAgenda,0)
    {\textbf{Research Agenda}\\
     {\color{black!72}
      \fontsize{5.18}{5.85}\selectfont
      Cross-cutting\\
      research priorities}\\
     {\fontsize{5.05}{5.75}\selectfont
      \textbf{Section VII}}};


\draw[umbrellaline]
    (topic.south) -- (\xAcq,\yUmbrella);

\draw[umbrellaline]
    (\xFound,\yUmbrella) -- (\xAgenda,\yUmbrella);

\draw[umbrellaline]
    (\xFound,\yUmbrella) -- (foundation.north);

\draw[umbrellaline]
    (\xArch,\yUmbrella) -- (architecture.north);

\draw[umbrellaline]
    (\xAcq,\yUmbrella) -- (acquisition.north);

\draw[umbrellaline]
    (\xEval,\yUmbrella) -- (evaluation.north);

\draw[umbrellaline]
    (\xAgenda,\yUmbrella) -- (agenda.north);


\draw[flowarrow]
    (foundation.east) -- (architecture.west);

\draw[flowarrow]
    (architecture.east) -- (acquisition.west);

\draw[flowarrow]
    (acquisition.east) -- (evaluation.west);

\draw[flowarrow]
    (evaluation.east) -- (agenda.west);


\node[itemFound] (f1) at (\xFound,\yOne)
    {Execution-substrate lens};

\node[itemFound] (f2) at (\xFound,\yTwo)
    {Workload-level\\
     boundary tests};

\node[itemFound] (f3) at (\xFound,\yThree)
    {Seven-dimensional\\
     competence profile};

\node[itemFound] (f4) at (\xFound,\yFour)
    {Adjacent-system\\
     comparators};

\coordinate (fRoute)
    at (\xFound,\yRoute);

\coordinate (fTrunkTop)
    at ($(f1.west)+(-0.20,0)$);

\coordinate (fTrunkBot)
    at ($(f4.west)+(-0.20,0)$);

\draw[foundline]
    (foundation.south) -- (fRoute);

\draw[foundline]
    (fRoute) -| (fTrunkTop);

\draw[foundline]
    (fTrunkTop) -- (fTrunkBot);

\foreach \n in {f1,f2,f3,f4}{
    \draw[foundline]
        (fTrunkTop |- \n.west) -- (\n.west);
}


\node[itemArch] (a1) at (\xArch,\yOne)
    {Overlapping design emphases};

\node[itemArch] (a2) at (\xArch,\yTwo)
    {Architectural\\
     responsibility layers};

\node[itemArch] (a3) at (\xArch,\yThree)
    {Architectural patterns\\
     and responsibility allocation};

\node[itemArch] (a4) at (\xArch,\yFour)
    {Trade-offs, evidence landscape,\\
     and attribution};

\coordinate (aRoute)
    at (\xArch,\yRoute);

\coordinate (aTrunkTop)
    at ($(a1.west)+(-0.20,0)$);

\coordinate (aTrunkBot)
    at ($(a4.west)+(-0.20,0)$);

\draw[archline]
    (architecture.south) -- (aRoute);

\draw[archline]
    (aRoute) -| (aTrunkTop);

\draw[archline]
    (aTrunkTop) -- (aTrunkBot);

\foreach \n in {a1,a2,a3,a4}{
    \draw[archline]
        (aTrunkTop |- \n.west) -- (\n.west);
}


\node[itemAcq] (q1) at (\xAcq,\yOne)
    {Collection environments\\
     and data sources};

\node[itemAcq] (q2) at (\xAcq,\yTwo)
    {Trajectory construction, filtering,\\
     and failure data};

\node[itemAcq] (q3) at (\xAcq,\yThree)
    {Learning signals and\\
     adaptation approaches};

\node[itemAcq] (q4) at (\xAcq,\yFour)
    {Transfer and\\
     coverage gaps};

\coordinate (qRoute)
    at (\xAcq,\yRoute);

\coordinate (qTrunkTop)
    at ($(q1.west)+(-0.20,0)$);

\coordinate (qTrunkBot)
    at ($(q4.west)+(-0.20,0)$);

\draw[acqline]
    (acquisition.south) -- (qRoute);

\draw[acqline]
    (qRoute) -| (qTrunkTop);

\draw[acqline]
    (qTrunkTop) -- (qTrunkBot);

\foreach \n in {q1,q2,q3,q4}{
    \draw[acqline]
        (qTrunkTop |- \n.west) -- (\n.west);
}


\node[itemEval] (e1) at (\xEval,\yOne)
    {Benchmark design emphases};

\node[itemEval] (e2) at (\xEval,\yTwo)
    {Evidence layers\\
     and process metrics};

\node[itemEval] (e3) at (\xEval,\yThree)
    {Protocol validity and\\
     competence observability};

\node[itemEval] (e4) at (\xEval,\yFour)
    {Measurement and\\
     attribution diagnostics};

\coordinate (eRoute)
    at (\xEval,\yRoute);

\coordinate (eTrunkTop)
    at ($(e1.west)+(-0.20,0)$);

\coordinate (eTrunkBot)
    at ($(e4.west)+(-0.20,0)$);

\draw[evalline]
    (evaluation.south) -- (eRoute);

\draw[evalline]
    (eRoute) -| (eTrunkTop);

\draw[evalline]
    (eTrunkTop) -- (eTrunkBot);

\foreach \n in {e1,e2,e3,e4}{
    \draw[evalline]
        (eTrunkTop |- \n.west) -- (\n.west);
}


\node[itemAgenda] (r1) at (\xAgenda,\yOne)
    {Cross-domain\\
     terminal competence};

\node[itemAgenda] (r2) at (\xAgenda,\yTwo)
    {Fresh and replayable\\
     process evaluation};

\node[itemAgenda] (r3) at (\xAgenda,\yThree)
    {Runtime governability\\
     and safety};

\node[itemAgenda] (r4) at (\xAgenda,\yFour)
    {Controlled model--harness\\
     attribution};

\coordinate (rRoute)
    at (\xAgenda,\yRoute);

\coordinate (rTrunkTop)
    at ($(r1.west)+(-0.20,0)$);

\coordinate (rTrunkBot)
    at ($(r4.west)+(-0.20,0)$);

\draw[agendaline]
    (agenda.south) -- (rRoute);

\draw[agendaline]
    (rRoute) -| (rTrunkTop);

\draw[agendaline]
    (rTrunkTop) -- (rTrunkBot);

\foreach \n in {r1,r2,r3,r4}{
    \draw[agendaline]
        (rTrunkTop |- \n.west) -- (\n.west);
}


\draw[
    black!22,
    line width=0.48pt
]
    (-1.48,\yRefTitle) -- (4.92,\yRefTitle);

\draw[
    black!22,
    line width=0.48pt
]
    (8.76,\yRefTitle) -- (15.16,\yRefTitle);

\node[
    align=center,
    text=black!62,
    font=\bfseries\fontsize{5.18}{5.90}\selectfont
] at (\xAcq,\yRefTitle)
    {Representative Literature};


\node[referenceCell] (cf) at (\xFound,\yReference)
    {SWE-agent~\cite{yang2024swe}\\
     Agentless~\cite{xia2024agentless}\\
     OSWorld~\cite{xie2024osworld}};

\node[referenceCell] (ca) at (\xArch,\yReference)
    {SWE-agent~\cite{yang2024swe}\\
     OpenHands~\cite{wang2025openhands}\\
     Meta-Harness~\cite{lee2026meta}};

\node[referenceCell] (cq) at (\xAcq,\yReference)
    {TerminalTraj~\cite{wu2026large}\\
     CLI-Gym~\cite{lin2026cli}\\
     AgentHER~\cite{ding2026agenther}};

\node[referenceCell] (ce) at (\xEval,\yReference)
    {Terminal-Bench~\cite{merrill2026terminal}\\
     SetupBench~\cite{arora2025setupbench}\\
     LongCLI-Bench~\cite{feng2026longcli}};

\node[referenceCell] (cr) at (\xAgenda,\yReference)
    {TerminalWorld~\cite{chu2026terminalworld}\\
     SWE-rebench~\cite{badertdinov2026swerebench}\\
     BashArena~\cite{kaufman2025basharena}\\
     Meta-Harness~\cite{lee2026meta}};


\draw[
    foundline,
    line width=0.82pt
]
    ($(cf.north west)+(0,0.04)$)
    --
    ($(cf.north east)+(0,0.04)$);

\draw[
    archline,
    line width=0.82pt
]
    ($(ca.north west)+(0,0.04)$)
    --
    ($(ca.north east)+(0,0.04)$);

\draw[
    acqline,
    line width=0.82pt
]
    ($(cq.north west)+(0,0.04)$)
    --
    ($(cq.north east)+(0,0.04)$);

\draw[
    evalline,
    line width=0.82pt
]
    ($(ce.north west)+(0,0.04)$)
    --
    ($(ce.north east)+(0,0.04)$);

\draw[
    agendaline,
    line width=0.82pt
]
    ($(cr.north west)+(0,0.04)$)
    --
    ($(cr.north east)+(0,0.04)$);

\end{tikzpicture}%
}

\caption{Analytical storyline of the survey.
The survey organizes the study of terminal agents around operational
foundation, system architecture, competence acquisition, evaluation and
diagnostics, and the resulting research agenda. Expanded branches summarize
the principal analytical content of Sections~II--VII. The column-aligned
reference strip provides representative literature anchors and boundary
comparators rather than one-to-one mappings to individual subtopics.}
\label{fig:article_organization}

\end{figure*}
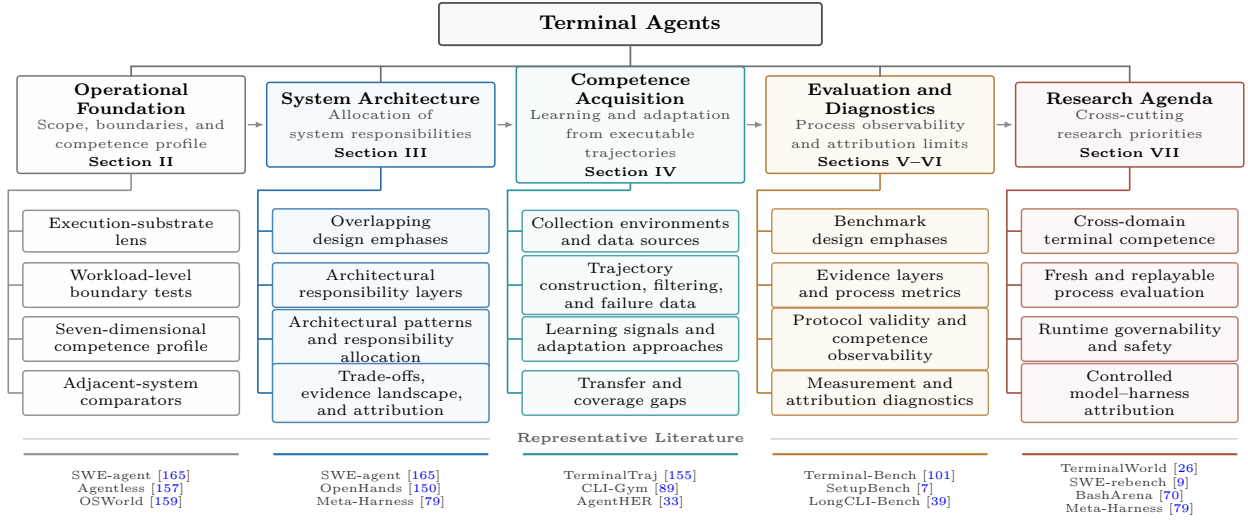

%% file: sections/2_background_terminal_agents.tex
\section{Background and Scope of Terminal Agents}
\label{sec:background_terminal_agents}

The execution-substrate lens requires both an operational boundary and a capability framework. This section establishes the workload-level scope and competence profile used throughout the survey.

\subsection{Terminal as a Command-Execution Agent Substrate}
\label{subsec:terminal_command_execution_substrate}

The terminal is a distinctive access layer for agentic execution because it combines compact textual interaction, compositional commands, persistent state, and broad operational reach~\cite{de2026terminal}. Through it, an agent can inspect directories, edit files, execute programs, install dependencies, launch processes, query logs, and access remote systems~\cite{bechard2026terminal}. Unlike GUI-centered environments, where state is inferred largely from visual layouts or interface events, terminal-mediated runtimes return stdout, stderr, exit codes, diffs, stack traces, logs, and process outputs in textual form, narrowing the gap between an LLM's token interface and the environment it controls~\cite{yang2024swe,lin2026cli}.

\begin{figure*}[!t]
\centering
\includegraphics[width=0.8\textwidth]{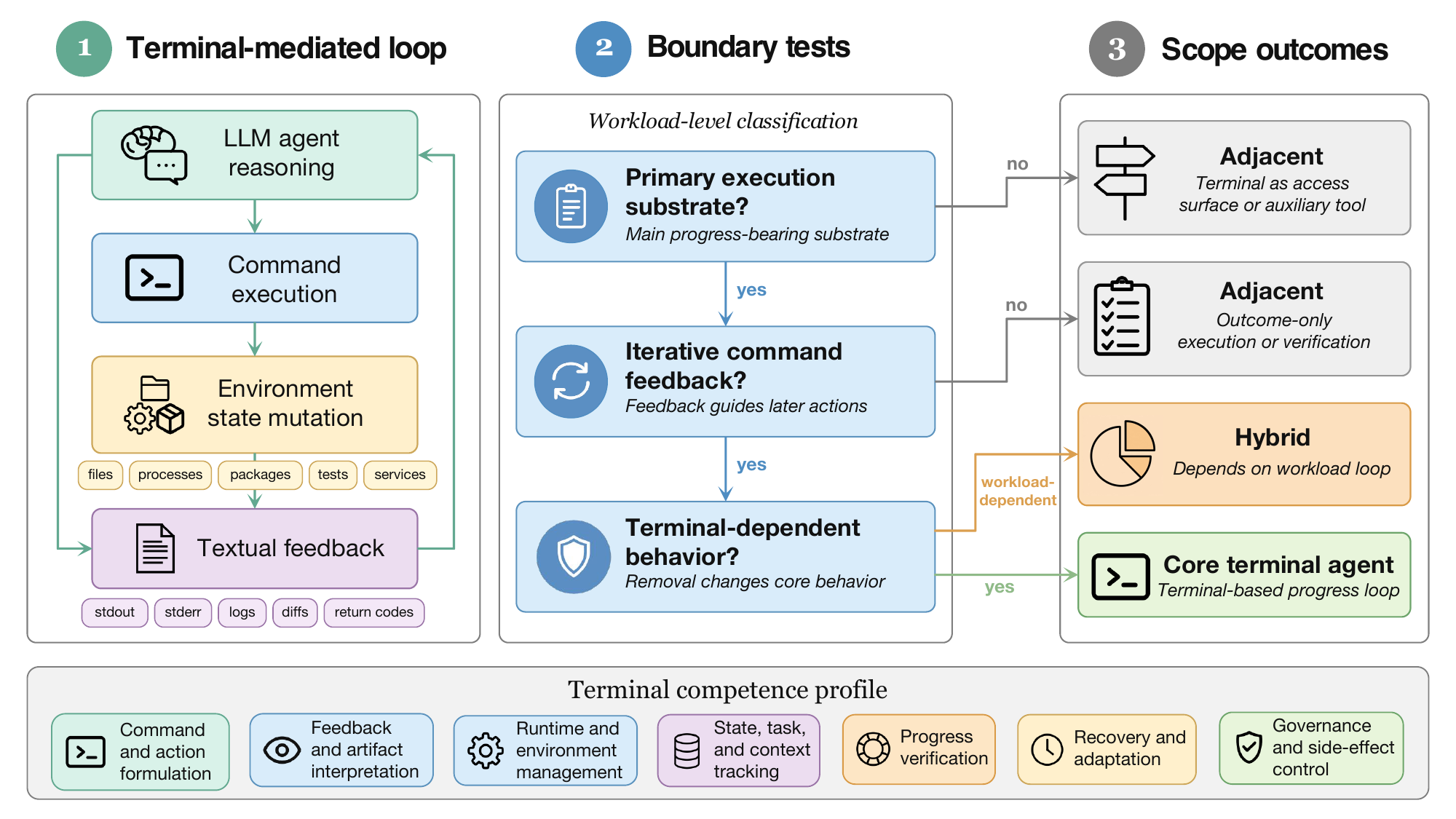}
\caption{Operational scope of terminal agents, linking terminal-mediated interaction to workload-level boundary tests, scope categories, and competence dimensions.}
\label{fig:terminal_agent_scope}
\end{figure*}

Four connected properties make terminal-mediated execution consequential: actions mutate persistent state; observations expose inspectable evidence; reruns and tests connect execution to verification; and commands, patches, and configuration edits form composable textual artifacts~\cite{de2026terminal,bechard2026terminal,yang2023intercode,wang2024executable,arora2025setupbench}. Together, these properties support a reasoning, execution, observation, and verification loop rather than merely a tool-call channel. Figure~\ref{fig:terminal_agent_scope} connects this loop to the boundary tests and competence dimensions introduced below.

Related terms describe different layers of this interaction. A \emph{terminal} is the textual input--output access layer; a \emph{shell} interprets commands; and a \emph{command-line tool} is a textual program invoked through a shell. A \emph{command-execution runtime} supplies the filesystem, processes, dependencies, permissions, and resource boundaries in which commands take effect. A \emph{harness} connects the model to this runtime by formatting observations, exposing actions, managing context, and enforcing execution policy. Terminal-mediated interaction therefore need not involve a visible terminal emulator or persistent pseudo-terminal; instead, command execution and the resulting textual and state evidence must carry task progress. A CLI-exposed product remains outside this scope when the CLI merely forwards requests to an otherwise non-terminal workflow.

\subsection{Substrate-Based Scope and Boundaries}
\label{subsec:substrate_based_scope_boundaries}

Building on this substrate view, we regard \emph{terminal agents} as systems whose dominant progress-bearing interaction is mediated by terminal command execution. The scope is substrate-based rather than surface-based: a system may appear through a CLI, IDE, web interface, or platform runtime, but is in scope only when task progress depends on command execution, textual feedback, and stateful environment interaction.

Occasional terminal use as an auxiliary tool is insufficient~\cite{parisi2022talm,schick2023toolformer}, as is one-shot code or patch generation without iterative execution and feedback. Terminal agents instead rely on an execution-grounded loop in which commands alter the environment and subsequent observations materially shape later decisions~\cite{yao2022react,wang2024executable,yang2024swe}. We operationalize this distinction with three workload-level tests:

\begin{itemize}[leftmargin=*,itemsep=1pt,topsep=2pt]
    \item \textbf{Primary execution substrate}: Terminal command execution is the workload's main means of progress.
    \item \textbf{Iterative command feedback}: Outputs, errors, logs, diffs, return codes, or state changes materially shape later actions.
    \item \textbf{Terminal dependence}: Removing terminal access would materially change the workload's core behavior.
\end{itemize}

``Dominant'' does not require every action to be a direct command; it means that task progress depends causally on terminal-mediated state changes and feedback. The tests apply to individual workloads rather than automatically to an entire product or platform. A hybrid platform may therefore be in scope for repository-repair or operations workloads when terminal execution drives state changes and subsequent decisions, but outside scope when another workload progresses mainly through a browser, GUI, or remote API.

Under these tests, SWE-agent is in scope when command feedback drives repository work~\cite{yang2024swe}, as are terminal-centric OpenHands workloads when terminal execution remains the principal locus of progress~\cite{wang2025openhands}. The tests exclude static patch generators without iterative execution~\cite{xia2024agentless}, GUI agents whose primary feedback is visual or DOM-based~\cite{xie2024osworld,zhou2024webarena}, and CLI-packaged assistants that use the terminal only as an access surface~\cite{theR1D2026shellgpt}. The same workload-level tests guide review-corpus screening.

\subsection{Terminal Competence as a Capability Profile}
\label{subsec:terminal_competence_profile}

Having established the workload boundary, we characterize terminal competence through major functional roles rather than low-level commands or domain skills. GUI environments such as OSWorld and AndroidWorld provide useful boundary comparators: they include environment state and executable verification, but visual interaction remains the progress-bearing substrate~\cite{xie2024osworld,rawles2025androidworld}. Across studies of systems, acquisition, and benchmarks, recurring behaviors and failures concern command selection, textual-feedback interpretation, environment setup and repair, state persistence, verification, recovery, long-horizon control, and execution governance~\cite{yang2024swe,wang2025openhands,arora2025setupbench,kuang2025process,merrill2026terminal,ding2026octobench}.

We group these responsibilities according to the object being controlled, the evidence required for the next decision, and the response needed when execution diverges from the task. This grouping separates runtime construction from persistent state tracking, verification from post-failure recovery, and authorization control from ordinary command execution. The resulting seven dimensions capture system-level responsibilities distributed across the model, interface, harness, runtime, and environment. They recur in interaction traces, system designs, training pipelines, and benchmark protocols, as summarized in Figure~\ref{fig:terminal_agent_scope}.

\begin{enumerate}[leftmargin=*,itemsep=1pt,topsep=2pt]
    \item \textbf{Command and action formulation}: translating goals, constraints, and state into executable commands, scripts, CLI calls, file edits, build/test/run actions, or structured terminal operations.

    \item \textbf{Feedback and artifact interpretation}: extracting task-relevant evidence from stdout, stderr, exit codes, logs, diffs, test outputs, stack traces, process signals, generated files, and workspace changes.

    \item \textbf{Runtime and environment management}: preparing, configuring, maintaining, and repairing dependencies, services, background processes, containers or virtual machines, remote machines, environment variables, resource limits, and related runtime constraints.

    \item \textbf{State, task, and context tracking}: maintaining environment state, task context, and interaction history across extended sessions, including filesystem and repository changes, packages, processes, configurations, prior commands, partial goals, verified facts, assumptions, and unresolved subproblems.

    \item \textbf{Progress verification}: designing and executing checks of intermediate validity, artifact trustworthiness, and completion conditions.

    \item \textbf{Recovery and adaptation}: diagnosing failures, revising hypotheses, replanning execution, mitigating harmful intermediate actions, and retrying from grounded evidence.

    \item \textbf{Governance and side-effect control}: respecting permissions, sandboxes, approval checkpoints, safety policies, credentials, resource limits, and restrictions on deletion, network access, privilege escalation, or external-system modification.
\end{enumerate}

These dimensions overlap with research on general tool use, software-engineering agents, and computer use, including work on tool selection, feedback, memory, planning, and safety~\cite{parisi2022talm,schick2023toolformer,yang2024swe,xie2024osworld,hu2025agents}. The profile organizes these concerns around the trace-observable obligations of terminal-mediated execution: commands mutate persistent runtime state, textual artifacts carry evidence across steps, verification occurs in the same substrate, and broad operational reach makes authorization and reversibility integral to competence. It thereby provides a common basis for comparing architecture, acquisition, and evaluation while preserving the distinctive demands of terminal-mediated execution.

The dimensions are analytically separable but operationally interdependent. Recovery depends on feedback interpretation, state tracking, and verification; runtime management depends on action formulation; and long-horizon persistence combines state tracking, verification, and recovery. The following sections use this profile to examine how these responsibilities are distributed across system architecture, acquired through executable interaction, and exposed by evaluation protocols. Section~\ref{sec:cross_cutting_insights} then illustrates selected measurement consequences of this synthesis.

%% file: sections/3_technical_evolution_system_architectures.tex
\section{Terminal-Agent System Design and Architectures}
\label{sec:terminal_agent_system_design_architectures}

Terminal competence emerges from interactions among the model, interface, runtime, control mechanisms, harness, and environment. This section traces how these components became explicit design objects, organizes their responsibilities into architectural layers and recurring patterns, and synthesizes the resulting trade-offs and attribution consequences.

\subsection{Shifts in Design Emphasis toward Terminal-Mediated Agency}
\label{subsec:architectural_evolution_terminal_mediated_agency}

\begin{figure*}[!t]
\centering
\includegraphics[width=0.8\textwidth]{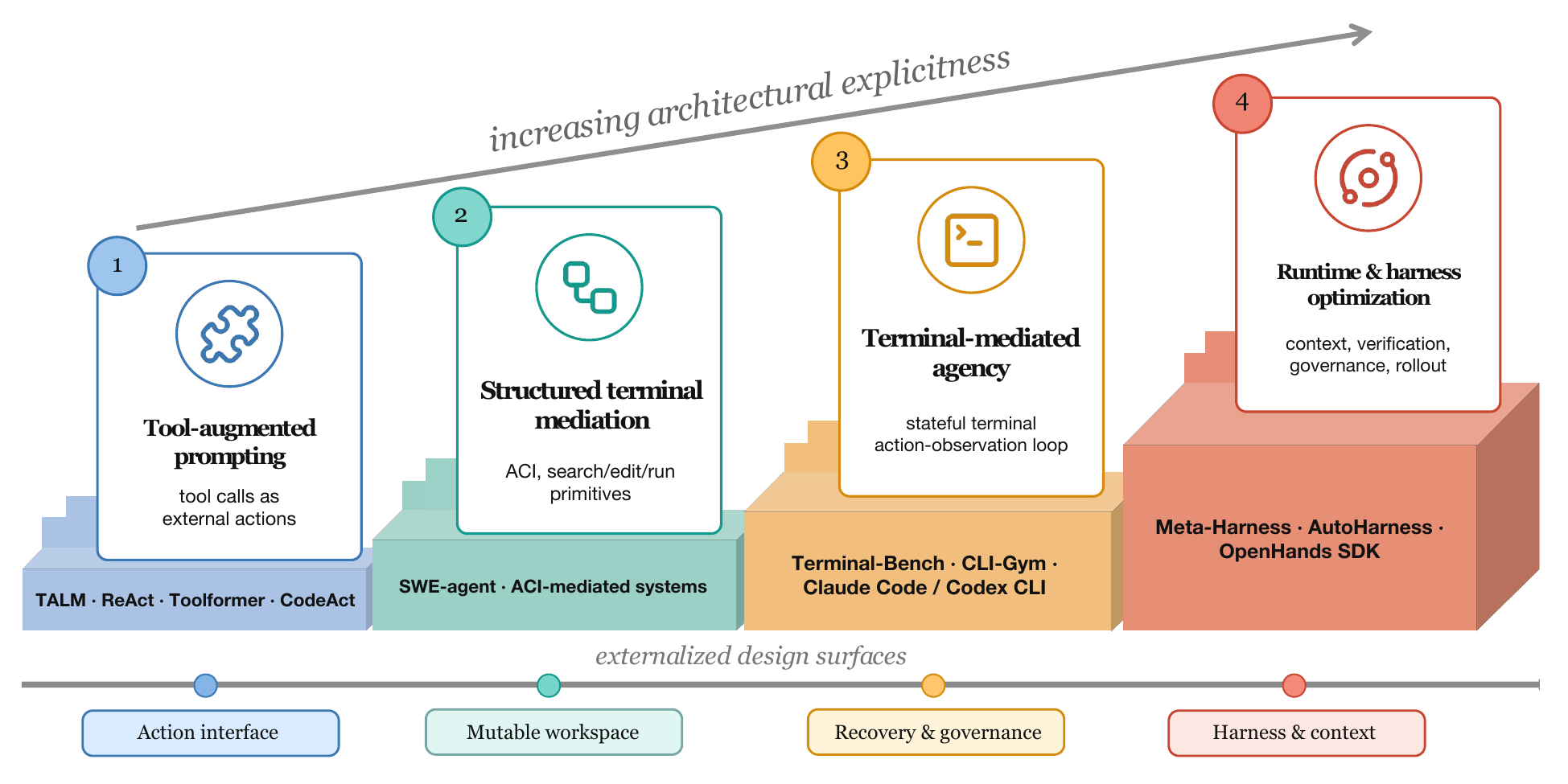}
\caption{Shifts in design emphasis toward terminal-mediated agency, from tool-augmented prompting to runtime- and harness-centered systems, with interfaces, workspaces, recovery, governance, and context management becoming increasingly explicit.}
\label{fig:technical_evolution_architecture}
\end{figure*}

Terminal-agent design can be understood through four overlapping shifts in emphasis: tool-augmented prompting, structured executable actions, terminal-mediated agency as a first-class target, and runtime- or harness-centered design. Rather than discrete generations, these shifts reflect the increasing treatment of action interfaces, mutable workspaces, recovery and governance policies, and harness-level context management as explicit design surfaces (Figure~\ref{fig:technical_evolution_architecture}).

\textbf{Shift 1: Tool-augmented prompting.}
TALM~\cite{parisi2022talm}, ReAct~\cite{yao2022react}, Toolformer~\cite{schick2023toolformer}, and CodeAct~\cite{wang2024executable} established observation-conditioned tool use while treating execution primarily as an external action channel rather than centering persistent terminal-mediated interaction.

\textbf{Shift 2: Structured executable actions.}
SWE-agent~\cite{yang2024swe} recasts repository navigation, editing, and execution as model-legible agent--computer interface primitives, moving from raw command access toward mediated interaction. OpenHands~\cite{wang2025openhands} extends this direction through a persistent platform runtime coordinating terminal, code, browser, and tool surfaces.

\textbf{Shift 3: Terminal-mediated agency as a first-class target.}
Terminal-Bench~\cite{merrill2026terminal}, CLI-Gym~\cite{lin2026cli}, Endless Terminals~\cite{gandhi2026endless}, and TermiGen~\cite{zhu2026termigen} treat terminal interaction as a training or evaluation distribution rather than a by-product of repository repair. Deployment studies document permission-gated command execution~\cite{bui2026building,bechard2026terminal}, while Claude Code~\cite{anthropic2025claudecode}, Codex CLI~\cite{openai2025codexcli}, Aider~\cite{gauthier2025aider}, and Gemini CLI~\cite{google2025geminicli} exhibit recurring patterns of command execution, workspace inspection, textual feedback, approval, and sandboxing.

\textbf{Shift 4: Runtime- and harness-centered design.}
Outer-loop elements such as context packaging, workspace persistence, verification, and control flow are increasingly optimized directly. Meta-Harness~\cite{lee2026meta}, AutoHarness~\cite{lou2026autoharness}, and Agentic Harness Engineering~\cite{lin2026agentic} treat context compaction, approval rules, observation shaping, and observability-guided harness evolution as performance-shaping variables. The OpenHands SDK~\cite{wang2025openhands} and extensible RL platforms~\cite{wang2025agentfly} make runtime behavior programmable, while collaborative frameworks extend orchestration across multiple agents or execution entities~\cite{foerster2026camels}. Interface, workspace, recovery, governance, and context management thus become explicit architectural components.

\subsection{Architectural Layers of Terminal Agents}
\label{subsec:architectural_layers_terminal_agents}

\begin{figure*}[!t]
    \centering
    \includegraphics[width=0.8\textwidth]{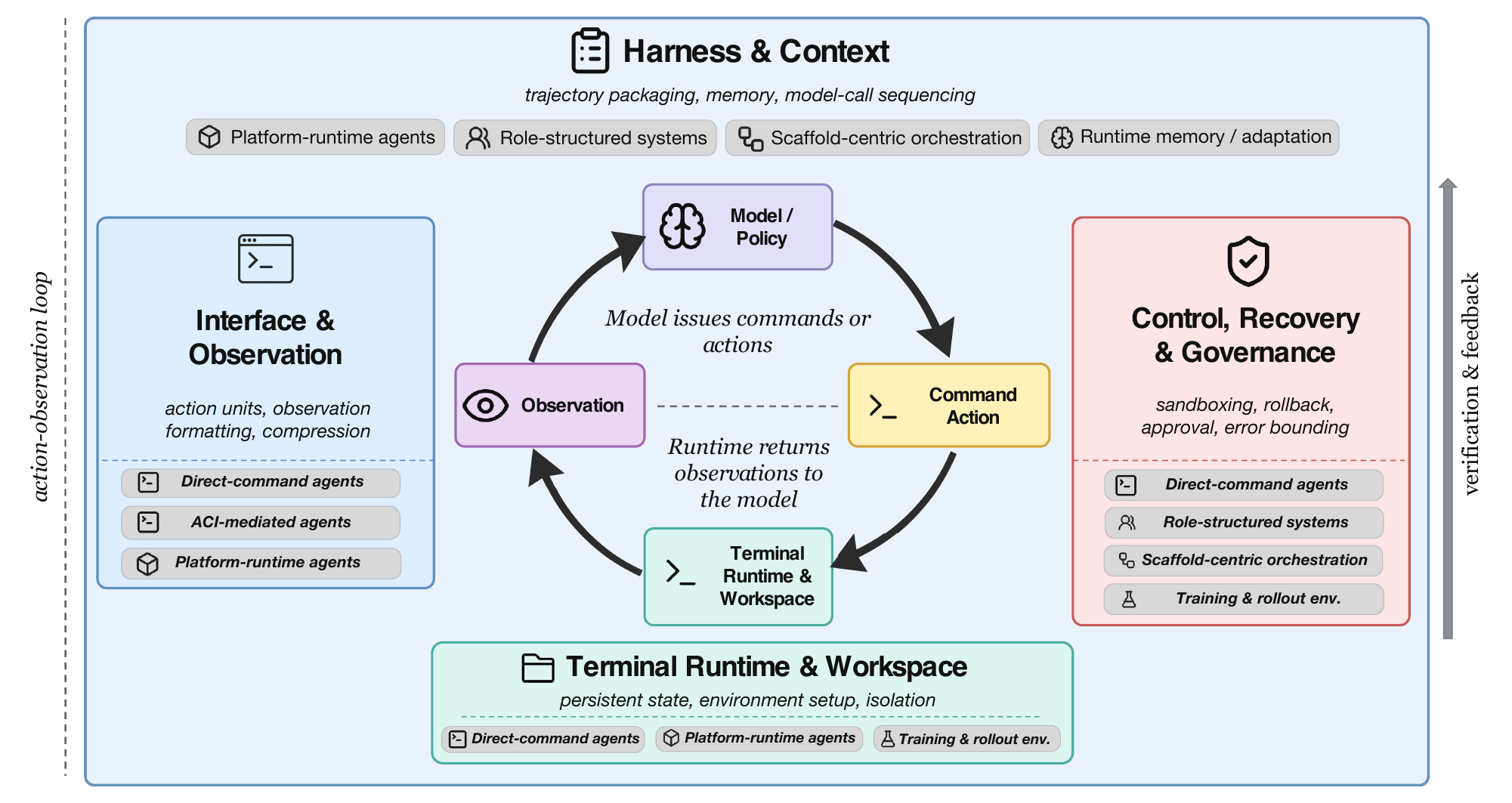}
    \caption{Layered architecture of terminal-agent systems. The central interaction loop is shaped by interface, runtime, verification, recovery, governance, and harness-level context mechanisms.}
    \label{fig:layered_architecture}
\end{figure*}

To compare these designs, we organize recurring responsibilities into four layers: interface and observation; runtime and workspace; control, verification, recovery, and governance; and harness and context. These layers locate where design choices enter the terminal-mediated loop and how they support the competence dimensions introduced in Section~\ref{sec:background_terminal_agents} (Figure~\ref{fig:layered_architecture}).

\textbf{Layer 1: Interface and observation.}
This layer specifies action units and feedback formats, ranging from direct commands to ACI primitives, runtime events, and workflow-stage interfaces. SWE-agent~\cite{yang2024swe} replaces unconstrained repository interaction with model-legible search, edit, and execution primitives. Observation design determines whether listings, traces, logs, exit codes, diffs, files, and workspace changes are exposed at useful granularity. In its reported setting, TACO reduces token use and improves accuracy through observational context compression~\cite{ren2026self}. Observation filtering and reinsertion further shape the scale and formatting of available feedback~\cite{de2026terminal}, while robustness across interface conditions remains open~\cite{rabinovich2025robustness}. This layer primarily supports dimensions 1 and 2.

\textbf{Layer 2: Runtime and workspace.}
Because commands mutate environment state, the runtime determines persistence, isolation, and whether later actions can build on earlier ones. OpenHands~\cite{wang2025openhands} coordinates terminal, code, browser, and tools over a shared workspace, whereas Terminal-Bench~\cite{merrill2026terminal} and CLI-Gym~\cite{lin2026cli} make controlled runtime environments central to evaluation. Persistent workspaces enable cumulative progress but can also propagate erroneous intermediate state. This layer governs dependencies, services, background processes, containers or virtual machines, resources, and workspace persistence, making it central to dimensions 3 and 4. Automated Docker image construction~\cite{zhang2026docksmith} further shows that workspace preparation can itself become an optimization target.

\textbf{Layer 3: Control, verification, recovery, and governance.}
Terminal access can modify filesystems, install packages, use credentials, and trigger external side effects. This layer structures checks, error bounding, recovery, and oversight through testing, sandboxing, rollback, approval, and decomposition. STRATUS uses role-delegated planning, execution, and review~\cite{chen2026stratus}; related principles appear in incident response~\cite{bilal2026large,lin2025ircopilot} and configuration-drift detection~\cite{abuzakuk2026riva}. AgentClick introduces skill-based checkpoints~\cite{zhuang2026agentclick}. OS-level resource management~\cite{mei2024aios,she2026agentrm}, verified deployment of generated Linux scheduling policies~\cite{zheng2025towards}, context-space access control~\cite{gong2025secure}, and reliable state management~\cite{thompson2026dualstatearchitecturereliablellm} further make permission, accountability, refusal, and side-effect control architectural concerns. This layer primarily supports dimensions 5 to 7.

\textbf{Layer 4: Harness and context.}
The harness packages history, formats prompts, manages context windows, and sequences model calls, determining what remains available across turns. Version-control-inspired methods offer alternatives to raw truncation~\cite{wu2025git}, and controlled comparisons show that context selection affects repository-level generation~\cite{le2025impacts}. Meta-Harness~\cite{lee2026meta} and AutoHarness~\cite{lou2026autoharness} report performance changes from outer-loop optimization, while multi-agent and asynchronous strategies broaden the orchestration space~\cite{benkovich2026agyn,geng2026effective}. This layer primarily supports dimension 4 and long-horizon persistence by preserving trajectory information, verified facts, unresolved subproblems, and prior actions.

Together, these layers show that terminal competence is distributed across the model, interface, runtime, control mechanisms, and harness. Long-horizon persistence therefore depends jointly on state and context tracking, runtime persistence, verification, and recovery.

A promising architectural direction is tighter state-aware coordination across these layers. Future systems could condition context management, verification, recovery, and governance on shared execution state and task progress, allowing observation, context allocation, verification, recovery, and execution policy to adapt coherently over long trajectories.

\subsection{Recurring Architectural Patterns and Responsibility Allocation}
\label{subsec:architecture_runtime_families}

The layers above locate architectural responsibilities, while recurring patterns describe how systems allocate them. These patterns are complementary rather than mutually exclusive and differ in action mediation, runtime persistence, control organization, context management, and rollout infrastructure. Direct-command access exposes broad command spaces behind permission or sandbox controls~\cite{anthropic2025claudecode,openai2025codexcli,gauthier2025aider,google2025geminicli}, whereas ACI mediation constrains search, edit, and execution through model-legible primitives~\cite{yang2024swe}. Platform runtimes coordinate persistent workspaces and multiple interaction surfaces~\cite{wang2025openhands}; role-structured systems separate planning, execution, and review~\cite{chen2026stratus,phan2024hyperagent}; and scaffold-centric systems manage context, observations, and control flow~\cite{lee2026meta,lou2026autoharness}. Runtime-memory methods compress or retrieve long-session state~\cite{sun2025scaling,ren2026self}, while terminal-native rollout environments provide scalable infrastructure for training and trajectory generation~\cite{lin2026cli,gandhi2026endless,zhu2026termigen,jain2025r2e}.

These patterns allocate responsibility differently. Direct access favors expressiveness but increases observation noise and rollback difficulty. Interface mediation favors reliability but may reduce cross-task generality. Persistent runtimes broaden the task surface while increasing dependence on the surrounding system, whereas context optimization supports long-horizon coherence while complicating attribution. Architectures are therefore better compared by how they allocate responsibility across the model, interface, runtime, control mechanisms, and harness than by assignment to a single family.

\subsection{Synthesis: Architectural Trade-offs and Attribution Consequences}
\label{subsec:architectural_tradeoffs}

These alternative responsibility allocations create three recurring architectural tensions and one attribution consequence.

\textbf{Expressiveness vs.\ recoverability.}
Raw or weakly mediated access expands the action space but produces noisier trajectories and harder rollback; ACI and scaffold constraints improve reliability yet may restrict other task families. This trade-off chiefly concerns dimensions 1 and 6, with verification supplying evidence for recovery.

\textbf{Generality vs.\ task discipline.}
Platform runtimes support heterogeneous terminal, code, and browser work but may provide less structured feedback. Role- or workflow-constrained systems strengthen local control at the cost of broader applicability. This tension primarily affects dimensions 3 and 4.

\textbf{Automation vs.\ inspectability.}
Deployable agents must keep commands, outputs, state changes, and interventions legible. Permission gates, approvals, and sandboxes improve auditability but add latency and interrupt autonomous execution. This tension centers on dimension 7, which remains among the least systematically addressed.

\textbf{Model capability vs.\ harness contribution.}
A further consequence is attribution difficulty: gains under optimized harnesses may arise from context management, observation shaping, retries, permission policy, or injected procedural knowledge rather than stronger model reasoning. Controlled skill injection produces task-dependent gains and can add substantial token overhead without improving pass rate~\cite{han2026swe}. Agentless likewise reports that static pipelines can rival interactive agents on some repository-repair tasks~\cite{xia2024agentless}. The complete model--harness--runtime configuration therefore remains relevant to system-level comparison, while component-level attribution requires separating these contributions.

\subsection{Evidence Landscape and Boundary Comparators}
\label{subsec:architecture_evidence_status}

The architectural literature spans established agent systems, controlled harness studies, deployment reports, and emerging training and operational settings. SWE-agent~\cite{yang2024swe} and OpenHands~\cite{wang2025openhands} provide broadly used system designs and evaluation settings, whereas Meta-Harness~\cite{lee2026meta} and AutoHarness~\cite{lou2026autoharness} examine harness optimization within individual studies. Deployment studies and commercial direct-command agents document recurring interface, permission, sandbox, and workflow patterns~\cite{bui2026building,bechard2026terminal,anthropic2025claudecode,openai2025codexcli,gauthier2025aider,google2025geminicli}. Training environments demonstrate scalable rollout generation, while evidence from live workflows and network operations is still emerging~\cite{garigipati2026beyond,nakamura2026helpful}.

Adjacent systems further clarify the architectural boundary. Agentless~\cite{xia2024agentless} represents static execution and verification pipelines, CGM~\cite{tao2026code} uses graph-structured control, OSWorld~\cite{xie2024osworld} grounds progress primarily in GUI interaction, and CLI-packaged assistants~\cite{theR1D2026shellgpt} use the terminal mainly as an access surface. These comparators distinguish progress-bearing terminal interaction from systems that share only selected architectural components.

Architectural choices determine executable actions, recorded observations and state changes, available interventions, and the failures entering a trajectory. They thereby shape both the supervision that acquisition pipelines can construct and the evidence that evaluation protocols can observe. Section~\ref{sec:terminal_competence_acquisition} examines how executable interactions become learning signals, while Section~\ref{sec:benchmarks_metrics_evaluation} examines how the resulting behavior becomes measurable evidence.

%% file: sections/4_terminal_competence_acquisition.tex
\section{Terminal Competence Acquisition and Adaptation}
\label{sec:terminal_competence_acquisition}

Building on the architectural account in Section~\ref{sec:terminal_agent_system_design_architectures}, competence acquisition concerns how executable interactions become learning and adaptation signals. Here, acquisition spans parameter learning and runtime adaptation. Training data inherit the action interface, harness, runtime, and governance conditions under which they are collected. Because commands mutate state, expose feedback and failures, and require continuation or repair decisions~\cite{pi2026data,gandhi2026endless,wu2026large,kang2026trace}, the relevant unit is a stateful trajectory containing actions, observations, state changes, verification, and recovery rather than an isolated prompt--response pair. This section follows the acquisition path from collection environments through trajectory construction and learning to runtime adaptation, then summarizes transfer and coverage gaps.

\begin{figure*}[!t]
    \centering
    \includegraphics[width=0.8\textwidth]{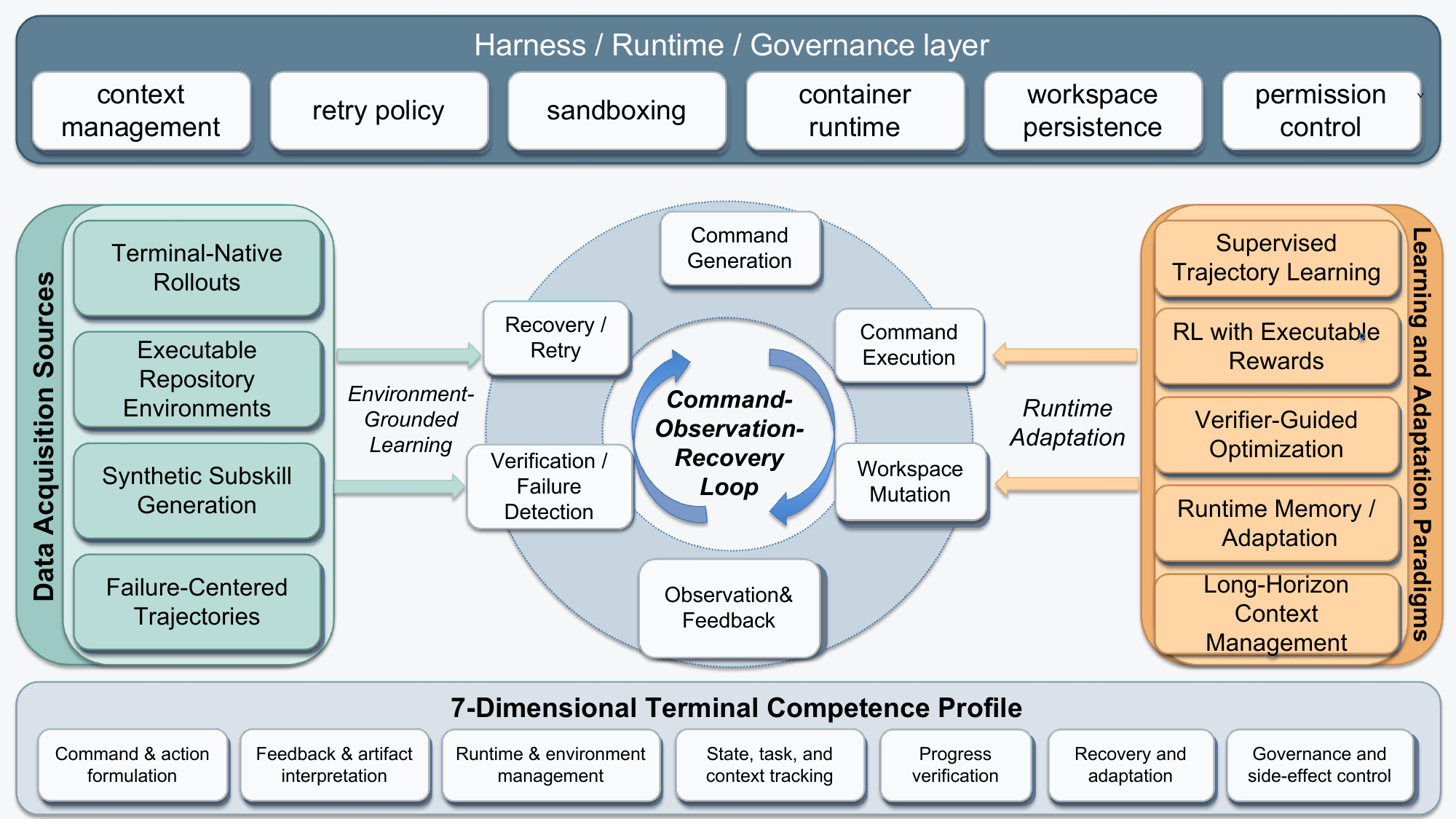}
    \caption{Terminal competence acquisition ecosystem. Data sources, interaction traces, runtime and governance mechanisms, and learning and adaptation approaches jointly shape the multidimensional competence profile.}
    \label{fig:terminal_competence_ecosystem}
\end{figure*}

Figure~\ref{fig:terminal_competence_ecosystem} connects acquisition sources, interaction and recovery traces, runtime and governance mechanisms, learning and adaptation approaches, and the resulting competence profile.

\subsection{Data Sources and Collection Settings for Terminal Competence}
\label{sec:competence_acquisition_sources}

Collection environments determine which actions, observations, state changes, and failures become available as learning evidence. Terminal-native rollouts expose command selection, output interpretation, and recovery~\cite{wu2026large,gandhi2026endless,lin2026cli,jain2025r2e}. Executable repository environments add navigation, editing, setup, and test-grounded verification, while coupling these behaviors to repository-specific reasoning~\cite{pan2024training,du2025swe,badertdinov2026swerebench,liang2026swe,guo2025swe}. Machine-learning engineering environments extend the loop to iterative experimentation and component refinement~\cite{qiang2026mle,nam2026mle}. Synthetic environments target sparse subskills but risk overfitting to generated patterns~\cite{zhu2026termigen,yang2026swe,meng2026calibforge}. Recent synthesis pipelines further scale executable data generation by jointly constructing instructions, environments, reference solutions, and verifiers through taxonomy-guided, evolutionary, or recursive procedures~\cite{hua2026cliuniverseverifiabletasksynthesis,shen2026seta,li2026recursivesynthesislonghorizonterminal}. Failure-centered corpora instead preserve diagnosis, rollback, and recovery that successful traces often omit~\cite{kang2026trace,ding2026agenther,zhang2026agentforesight}.

Environment structure also determines which competence dimensions are exposed for supervision. Short rollouts primarily expose action formulation and feedback interpretation; setup tasks add runtime management; long trajectories stress state and context tracking; and executable checks support verification. Retained failures expose diagnosis and recovery, while governance requires settings in which authorization, containment, and side effects are visible.

\subsection{Trajectory Construction, Filtering, and Failure Data}
\label{sec:trajectory_construction_failure_data}

A rollout becomes usable training data through selection, validation, filtering, and relabeling. These operations shape both data quality and the behaviors preserved for learning~\cite{wu2026large,xu2026cleaner,ding2026agenther,pi2026data,zhu2026termigen,kang2026trace}. Dockerized generation integrates task adaptation, synthetic generation, rollout collection, filtering, and decontamination~\cite{wu2026large}. CLEANER filters trajectories for reinforcement learning~\cite{xu2026cleaner}, while AgentHER validates and relabels them through hindsight replay~\cite{ding2026agenther}. Trajectory utility also depends on preserved interaction structure: TerminalLego reports stronger training signals from environment-grounded inspect--act--verify trajectories than from teacher success alone in its setting~\cite{yang2026makesinteractiontrajectorieseffective}. Other mechanisms alter what remains visible during learning: progressive code masking varies access to prior code~\cite{kim2025reproduction}; Nemotron-Terminal constructs terminal-oriented training data~\cite{pi2026data}; Live-SWE-agent studies online self-evolution~\cite{xia2025live}; and long-context multi-turn reinforcement learning retains extended interactions~\cite{golubev2025training}.

Recoverable failures remain underrepresented. Failed installations, version conflicts, invalid environment assumptions, rollback decisions, and dead-end repairs expose recovery more directly than final successful commands~\cite{kang2026trace,ding2026agenther,zhang2026agentforesight,zhao2026debugging,zhu2025llm}. Yet successful-trace filtering can remove the interactions needed to learn diagnosis and adaptation~\cite{xu2026cleaner,ding2026agenther,zhu2025llm}. Retaining misdiagnosis, rollback, and repair therefore provides direct supervision for recovery under observed failure modes~\cite{kang2026trace,ding2026agenther,zhao2026debugging}.

\subsection{Learning Signals and Adaptation Approaches}
\label{sec:learning_signals_post_training}

Constructed trajectories support complementary levers at three levels: data construction, parameter optimization, and runtime adaptation. These levers can be combined within a single acquisition pipeline. Table~\ref{tab:acquisition_paradigms} compares their primary signal, capability target, blind spot, and principal risk.

\begin{table*}[!t]
\caption{Complementary approaches to terminal competence acquisition and adaptation.}
\label{tab:acquisition_paradigms}
\centering
\footnotesize
\setlength{\tabcolsep}{2.5pt}
\renewcommand{\arraystretch}{1.03}

\begin{tabularx}{\textwidth}{
@{}
>{\raggedright\arraybackslash}p{3.75cm}
>{\hsize=0.98\hsize\linewidth=\hsize
  \raggedright\arraybackslash}X
>{\hsize=1.02\hsize\linewidth=\hsize
  \raggedright\arraybackslash}X
>{\hsize=1.08\hsize\linewidth=\hsize
  \raggedright\arraybackslash}X
>{\hsize=0.92\hsize\linewidth=\hsize
  \raggedright\arraybackslash}X
@{}}
\toprule
Approach
& Primary signal
& Capability target
& Blind spot
& Key risk \\
\midrule

SFT on successful traces
& Successful interaction traces
& Common commands, repository navigation, and standard workflows
& Recovery, rollback, and diagnosis after wrong assumptions
& Clean-trace overfit; absent failure paths \\

\mbox{RL with environment rewards}
& Executable success, verifier reward, and test outcome
& Outcome-driven exploration and completion
& Process quality and safe intermediate behavior
& Sparse-reward hacking; benchmark overfit \\

Process-aware / verifier-guided
& Step judgments and verifier rankings
& Diagnosis, action selection, and recovery decisions
& Open-domain transfer
& Verifier bias and shifted failures \\

Synthetic task generation
& Generated tasks with controlled verification
& Rare commands, setup patterns, and targeted subskills
& Live realism and environment drift
& Synthetic heuristics; weak transfer \\

\mbox{Failure-conditioned training}
& Failed or repaired traces; hindsight relabeling
& Recovery, diagnosis, and early failure detection
& Generalization across inconsistent failure taxonomies
& Noisy labels; repair-loop overfit \\

Runtime memory / context adaptation
& Session history, summaries, and retrieved experience
& Long-horizon persistence and workflow reuse
& Correction under incorrect memory
& Compression loss; memory contamination \\

\bottomrule
\end{tabularx}
\end{table*}

At the parameter-optimization level, \textbf{SFT on successful traces} teaches common commands, repository workflows, and setup patterns. Nemotron-Terminal~\cite{pi2026data} uses terminal-oriented data engineering, while SWE-Gym~\cite{pan2024training} and SWE-Dev~\cite{du2025swe} provide executable software trajectories. Their emphasis on successful interactions, however, offers limited supervision for diagnosis, rollback, and recovery after incorrect assumptions~\cite{yang2025kimi,zeng2026davinci}. \textbf{RL with executable rewards} aligns behavior with task completion through environments and verifiers, as in Endless Terminals~\cite{gandhi2026endless}, ECHO~\cite{shrivastava2026echo}, SWE-Master~\cite{song2026swe}, SWE-Gym~\cite{pan2024training}, and Tmax~\cite{ivison2026tmaxsimplerecipeterminal}, but sparse rewards may reinforce brittle or unsafe behavior. \textbf{Process-aware and verifier-guided optimization} instead supervises intermediate decisions through judgments, rankings, or hindsight validation~\cite{wei2026swe,ding2026agenther}. AgentHER~\cite{ding2026agenther} is especially relevant because terminal failures often appear early through stderr, logs, failed tests, or inconsistent state.

Data-oriented and runtime approaches complement parameter optimization. \textbf{Synthetic task generation} expands coverage of rare commands, dependency search, repair, localization, and environment construction through verifiable tasks~\cite{zhu2026termigen,yang2026swe,meng2026calibforge}, but may teach synthetic regularities rather than reusable competence. \textbf{Failure-conditioned training} uses repaired or failed traces from TRACE~\cite{kang2026trace}, AgentHER~\cite{ding2026agenther}, and AgentForesight~\cite{zhang2026agentforesight} for diagnosis, early failure prediction, hindsight relabeling, and recovery; its main obstacles are noisy labels, inconsistent taxonomies, and repair-loop overfit. \textbf{Runtime memory and context adaptation} extends acquisition into online behavior: Memento~\cite{zhou2025memento}, Context-Folding~\cite{sun2025scaling}, and TACO~\cite{ren2026self} retain history, summaries, or compressed context across long interactions, while risking persistence of incorrect assumptions.

These approaches supervise different parts of an interaction history: successful traces teach common workflows, executable rewards favor completion, verifiers and hindsight expose intermediate decisions, synthetic tasks broaden coverage, failure-conditioned data supports recovery, and runtime memory sustains long-horizon behavior. Their shared challenge is to combine these signals without overfitting to benchmark tasks, generated environments, or harness-specific rewards.

\subsection{Emerging Practices and Remaining Gaps}
\label{sec:acquisition_emerging_gaps}

Recent systems increasingly treat acquisition as an end-to-end pipeline in which sourcing, rollout generation, filtering, replay, and curriculum design are distinct levers~\cite{pi2026data,pan2024training,wu2026large}. Such pipelines increasingly retain setup attempts, outputs, state changes, failed commands, verifier feedback, and recovery decisions, yet current corpora remain concentrated on successful repository-centric traces, with limited coverage of failed setup, long repair loops, environment drift, and unsafe intermediate actions~\cite{kang2026trace,xu2026cleaner,ding2026agenther}.

Transfer remains a central unresolved issue. Repository-repair training may produce task-specific heuristics rather than general terminal competence~\cite{xia2024agentless}, motivating work on action, observation, and reward sequences that encode reusable skills across environments~\cite{li2026agencybench}. Work on externalization~\cite{zhou2026externalization} and semantics-aware repair~\cite{pabba2025semagent} further suggests that transfer depends on how explicitly trajectories preserve intermediate reasoning and execution evidence. Cross-domain workflows, environment drift, failed setup, unsafe intermediate actions, and long repair loops remain weakly represented.

These gaps make evaluation evidence essential for determining which aspects of terminal competence are actually acquired. Final success alone does not reveal whether an agent preserved state, recovered from failure, verified completion, or respected execution constraints. Section~\ref{sec:benchmarks_metrics_evaluation} therefore examines which benchmark families and evidence layers make these behaviors observable, helping distinguish transferable terminal competence from adaptation to particular repositories, harnesses, environments, or reward channels.

%% file: sections/5_benchmarks_metrics_evaluation.tex
\section{Benchmarks, Metrics, and Evaluation}
\label{sec:benchmarks_metrics_evaluation}

Evaluation determines which aspects of the architectures and acquisition pipelines discussed above become observable and which remain hidden behind final outcomes. Terminal-agent evaluation spans software engineering, tool use, and environment-coupled interaction, making it important to distinguish \emph{terminal competence} from \emph{repository repair competence}~\cite{jimenez2024swe,yang2024swe,arora2025setupbench,merrill2026terminal}. The two overlap but are not equivalent.

We organize the literature along four analytical axes: benchmark design emphases characterize task settings and evaluation targets; evidence layers specify what is recorded; the coverage map estimates which competence dimensions become observable; and protocol validity determines how scores can be interpreted across evaluation settings. Together, these axes separate what a benchmark asks agents to do, what evidence it records, and what conclusions its protocol supports.

Table~\ref{tab:benchmark_families} groups non-exclusive benchmark design emphases by their primary evaluation focus and principal blind spot. They are descriptive rather than ranked because their tasks, scoring signals, and execution infrastructure differ.

\begin{table*}[!t]
\caption{Benchmark design emphases by primary evaluation focus and principal blind spot.}
\label{tab:benchmark_families}

\centering
\footnotesize
\setlength{\tabcolsep}{2.5pt}
\renewcommand{\arraystretch}{0.94}

\begin{tabularx}{\textwidth}{
@{}
>{\raggedright\arraybackslash}p{3.15cm}
>{\raggedright\arraybackslash}p{3.75cm}
>{\raggedright\arraybackslash}p{4.55cm}
>{\raggedright\arraybackslash}X
@{}
}

\toprule

\textbf{Emphasis}
& \textbf{Primary evaluation focus}
& \textbf{Representative examples}
& \textbf{Principal blind spot}
\\

\midrule

Repository repair
& Issue resolution under tests
& \mbox{SWE-bench} and \mbox{SWE-PolyBench}%
  ~\cite{jimenez2024swe,rashid2025swe}
& Conflates repository repair with terminal competence; command use is not directly measured
\\

\mbox{CLI/terminal-centered}
& Commands, output interpretation, and CLI workflows
& \mbox{Terminal-Bench}, \mbox{TerminalWorld}, and \mbox{LongCLI-Bench}%
  ~\cite{merrill2026terminal,chu2026terminalworld,feng2026longcli}
& Mixes terminal-native and repository-mediated tasks, leaving transfer unclear
\\

Setup
& Environment setup and dependency resolution
& \mbox{SetupBench}~\cite{arora2025setupbench}
& Often isolated from end-to-end workflows
\\

Process
& Intermediate behavior and scaffold compliance
& \mbox{OctoBench}, \mbox{ProcBench}, and \mbox{AppWorld}%
  ~\cite{ding2026octobench,he2026procbench,trivedi2024appworld}
& Lacks standardized process scoring
\\

Long horizon
& Persistence, drift, and temporal coherence
& \mbox{SWE-Bench Pro}, \mbox{LoCoEval}, and \mbox{LifelongAgentBench}%
  ~\cite{deng2025swe,liu2026scalable,zheng2025lifelongagentbench}
& Costly and difficult to reproduce at scale
\\

\mbox{Safety/governance}
& Permissions, containment, and privileged commands
& \mbox{BashArena}, \mbox{ClawSafety}, and \mbox{AgentHazard}%
  ~\cite{kaufman2025basharena,wei2026clawsafety,feng2026agenthazard}
& Immature protocols; task success may conceal harmful actions
\\

Production
& Distribution match and deployment realism
& \mbox{ProdCodeBench}~\cite{jha2026reap}
& Limited public access and domain coverage
\\

\bottomrule
\end{tabularx}
\end{table*}

Production-derived evaluation cuts across these emphases by improving distribution match without isolating a single terminal skill~\cite{jha2026reap,jha2025itbench}. Repository-scale deployment studies add evidence about adoption, quality, security-relevant changes, and task-conditioned acceptance~\cite{robbes2026agentic,agarwal2026ai,siddiq2026security,pinna2026comparing}, but provide limited evidence about individual dimensions of terminal competence.

\subsection{From Repository-Level Evaluation to Terminal-Native Benchmarks}
\label{subsec:repo_to_terminal_benchmarks}

SWE-bench~\cite{jimenez2024swe} established repository issue resolution with executable verification as a dominant evaluation paradigm. Successors expanded language coverage~\cite{rashid2025swe,zan2026multi}, project scale~\cite{liang2026swe}, temporal validity~\cite{sun2026atime,badertdinov2026swerebench}, production realism~\cite{jha2026reap}, and multilingual agentic evaluation~\cite{adamenko2025swe}. SWE-Hub integrates environment construction, task synthesis, and executable validation in a scalable production pipeline~\cite{zeng2026swe}. Claw-SWE-Bench adds multilingual repository tasks and a Lite subset suited to matched system comparisons~\cite{zheng2026claw}. Section~\ref{sec:cross_cutting_insights} pairs Claw-SWE-Bench Lite with SWE-bench Lite: the former broadens repository and language diversity, while the latter provides a widely used compatibility anchor. Both remain repository-repair protocols with indirect coverage of broader terminal competence.

Terminal-native and CLI-centered benchmarks instead make command-mediated interaction part of the task definition~\cite{merrill2026terminal,chu2026terminalworld,feng2026longcli,ni2026gittaskbench,ding2026wildclawbench,lai2026clawforge}. They directly expose command use, output interpretation, state inspection, and workflow persistence. TUA-Bench extends this design beyond technical and programming-centric workflows to routine digital activities and scientific and engineering work conducted through a terminal~\cite{chen2026tuabenchbenchmarkgeneralpurposeterminaluse}. CLI-Gym~\cite{lin2026cli} supports both training and evaluation, while InterCode~\cite{yang2023intercode} anticipated this direction through interactive coding with execution feedback. This group remains heterogeneous: tasks range from terminal-native workflows to repository-mediated work, WildClawBench reports an 18-point harness-conditioned gap~\cite{ding2026wildclawbench}, and ClawForge identifies proactive state inspection as a strong discriminator in its setting~\cite{lai2026clawforge}. Transfer across these task types remains insufficiently established.

\subsection{Process-Aware, Environment-Aware, and Long-Horizon Evaluation}
\label{subsec:process_environment_long_horizon_eval}

Evaluation is expanding from measuring only \emph{what} agents produce to examining \emph{how} they interact with mutable environments. \textbf{Process-aware evaluation} separates final success from intermediate quality and scaffold compliance. OctoBench~\cite{ding2026octobench} and debug-gym~\cite{yuan2025debug} add step-level assessment and defect ontologies; ProcBench~\cite{he2026procbench} and AgentEval~\cite{guo2026agenteval} assess error propagation and control preservation; and ToolSandbox~\cite{lu2025toolsandbox}, AppWorld~\cite{trivedi2024appworld}, and ASTRA-Bench~\cite{xiu2026astra} extend stateful tool use and action planning beyond repository repair. These efforts expose intermediate behavior, although no shared process-scoring standard has emerged.

\textbf{Environment-aware evaluation} treats setup and dependency resolution as task components rather than pre-task overhead. SetupBench isolates environment bootstrap~\cite{arora2025setupbench}, while process-level configuration studies identify setup defects that binary success misses~\cite{kuang2025process}. Environment setup thereby becomes an explicit component of terminal competence and end-to-end evaluation.

\textbf{Long-horizon evaluation} reveals failures hidden by short tasks. In its reported setting, SWE-EVO~\cite{thai2025swe} reports a drop from 65--73\% on SWE-Bench Verified to 21--25\% on multi-file evolution. SWE-Bench Pro~\cite{deng2025swe}, LoCoEval~\cite{liu2026scalable}, SlopCodeBench~\cite{orlanski2026slopcodebench}, Spec Emerges~\cite{yan2026specification}, ProjDevBench~\cite{lu2026projdevbench}, and RepoMod-Bench~\cite{li2026repomod} similarly expose degradation under repeated editing. Long-Horizon-Terminal-Bench adds dense subtask rewards and partial-credit grading to terminal workflows that require sustained execution~\cite{li2026longhorizonterminalbenchtestinglimitsagents}. NL2Repo-Bench~\cite{ding2025nl2repo} targets repository generation, AgencyBench~\cite{li2026agencybench} stresses 1M-token contexts, LifelongAgentBench~\cite{zheng2025lifelongagentbench} evaluates lifelong behavior, and WildClawBench~\cite{ding2026wildclawbench} covers deployment scenarios. These settings extend evaluation from local completion to sustained state coherence, progress tracking, and alignment over time.

\subsection{Protocol Validity and Harness Effects}
\label{subsec:protocol_validity_harness_effects}

Evaluation scores reflect a coupled configuration of the model, task set, and protocol choices governing observation, execution, retries, memory, and runtime conditions~\cite{yang2024swe,lee2026meta,lou2026autoharness,ge2026agent}. Three factors are especially important for interpreting comparisons across evaluation settings.

\textbf{Contamination and temporal validity.}
Static task pools are vulnerable to memorization, prompt leakage, stale distributions, and public exposure of issues, patches, tests, or discussions. AgentBench~\cite{liu2024agentbench} established broader agent-evaluation protocols, while mutation and rebenchmarking improve freshness~\cite{garg2025saving,badertdinov2026swerebench,badertdinov2026swerebenchv2}. LiveSQLBench~\cite{bird2024livesqlbench} extends dynamic evaluation to database tasks, ACE-Bench~\cite{yang2026ace} varies difficulty and horizon, and contamination detection distinguishes recall from reasoning~\cite{song2026cross}. Task construction is equally important: an audit reports that 16\% of tasks across five terminal-agent benchmarks are reward-hackable~\cite{zhong2026hardening}. These developments make temporal freshness, contamination control, task regeneration, and resistance to reward hacking central to protocol validity.

\textbf{Harness-mediated variance.}
The harness is part of the evaluated configuration. Observation formatting, approval, sandboxing, retry limits, context truncation, wrappers, and recovery affordances can alter performance under a fixed model. Controlled comparisons across CLI and MCP interfaces make these effects explicit~\cite{alierforment2026scaffolding}; Meta-Harness~\cite{lee2026meta} and AutoHarness~\cite{lou2026autoharness} optimize the outer loop directly, while Agent Psychometrics~\cite{ge2026agent} separates LLM and scaffold ability through item response theory. Controlled terminal evaluations further show that harness choice changes token efficiency and failure profiles under fixed models, while AgentMeter jointly evaluates task quality, budget sensitivity, and LM-CLI matching~\cite{vats2026scaffoldeffectcodingagents,chi2026matchingmattersfairqualityefficiency}. Model comparisons therefore depend on the harness conditions under which behavior is produced.

\textbf{Environment and budget comparability.}
Container images, dependency caches, network access, timeouts, filesystem persistence, and tool permissions alter both success and failure modes~\cite{arora2025setupbench,kuang2025process,merrill2026terminal}. Production-derived tasks reduce distribution mismatch but may trade breadth and controlled conditions for realism~\cite{jha2026reap}. Retry budgets, model and verifier calls, trajectory limits, and wall-clock timeouts further determine the search space available to an agent.

Together, these factors make the model, harness, observation and action interfaces, execution environment, permission policies, and execution budget part of the evaluated configuration~\cite{yang2024swe,lee2026meta,lou2026autoharness,ge2026agent}.

\subsection{Evaluation Layers and Metrics Beyond Binary Correctness}
\label{subsec:metric_design_beyond_binary}

Resolution and pass rates remain leaderboard anchors but capture only final correctness. Complementary evidence includes behavioral analysis beyond resolution rate~\cite{mehtiyev2026beyond}, syntax-aware structure in SWE-PolyBench~\cite{rashid2025swe}, checklist-based process scores in OctoBench~\cite{ding2026octobench}, long-horizon drift and faithfulness~\cite{yan2026specification,orlanski2026slopcodebench}, and multidimensional enterprise assessment~\cite{chandwani2026beyond}. Relevant process indicators include \emph{command economy}, relating command number and complexity to task scope; \emph{recovery}, recording actions after failure; \emph{diagnostic quality}, assessing whether causes are identified before repair; \emph{state tracking}, checking consistency with filesystem, package, process, and repository state; and \emph{governance violations}, capturing unsafe commands, permission bypass, sandbox escape, or unapproved external effects.

These signals form a layered evidence stack. Outcome evidence records task completion; process evidence describes execution; environment evidence establishes runtime validity; trace evidence supports inspection and replay; and governance evidence records permissions, containment, and side effects. Freshness is not a behavioral layer but a cross-cutting condition of evaluation validity. Deployment-oriented multi-signal evaluation follows the same motivation~\cite{gao2026agentpulse}. Outcome metrics are comparatively standardized, whereas the remaining layers use heterogeneous definitions and protocols~\cite{ding2026octobench,he2026procbench,bouzenia2025understanding,ceka2025understanding,kaufman2025basharena,wei2026clawsafety}.

\subsection{Benchmark Observability of Terminal Competence Dimensions}
\label{subsec:benchmark_competence_coverage}

Table~\ref{tab:benchmark_families} characterizes benchmark design emphases, while Table~\ref{tab:benchmark_capability_coverage} maps representative groups to the seven competence dimensions and treats freshness separately as evaluation validity. The coding indicates which constructs each group makes observable rather than the strength of their measurement. \textbf{Explicit} denotes a targeted or scored construct, \textbf{Trace-visible} a required but unscored construct, \textbf{Incidental} a construct that may arise without being targeted, and \textbf{Not observable} little basis for observation or scoring. The rows include core terminal-agent benchmarks and terminal-relevant boundary comparators; grouped rows reflect their shared dominant design intent.

\begin{table*}[!t]
\caption{Qualitative coding of terminal-competence observability and evaluation freshness. E=Explicit, T=Trace-visible, I=Incidental, and --=Not observable. Freshness concerns evaluation validity rather than competence.}
\label{tab:benchmark_capability_coverage}

\centering
\footnotesize
\setlength{\tabcolsep}{2.6pt}
\renewcommand{\arraystretch}{0.96}

\begin{tabularx}{\textwidth}{
@{}
>{\raggedright\arraybackslash}X
*{7}{>{\centering\arraybackslash}p{1.20cm}}
>{\centering\arraybackslash}p{1.05cm}
@{}
}

\toprule

\textbf{Benchmark}
& \multicolumn{7}{c}{\textbf{Terminal-competence dimensions}}
& \makecell[c]{\textbf{Eval.}\\\textbf{validity}}
\\

\cmidrule(lr){2-8}
\cmidrule(l){9-9}

& \makecell[c]{\textbf{Action}\\\textbf{form.}}
& \makecell[c]{\textbf{Feedback}\\\textbf{interp.}}
& \makecell[c]{\textbf{Runtime}\\\textbf{mgmt.}}
& \makecell[c]{\textbf{State}\\\textbf{tracking}}
& \textbf{Verify}
& \textbf{Recover}
& \textbf{Govern}
& \textbf{Fresh.}
\\

\midrule

\mbox{SWE-bench / SWE-PolyBench~\cite{jimenez2024swe,rashid2025swe}}
& I & T & I & T & T & I & -- & -- \\

Claw-SWE-Bench~\cite{zheng2026claw}
& I & T & I & T & T & I & -- & I \\

\mbox{Terminal-Bench / TerminalWorld~\cite{merrill2026terminal,chu2026terminalworld}}
& E & E & T & T & T & T & -- & I \\

SetupBench~\cite{arora2025setupbench}
& E & E & E & T & T & T & -- & -- \\

\mbox{LongCLI-Bench / GitTaskBench~\cite{feng2026longcli,ni2026gittaskbench}}
& E & E & I & E & T & I & -- & -- \\

CLI-Gym~\cite{lin2026cli}
& E & E & T & T & T & T & -- & -- \\

\mbox{debug-gym / OctoBench~\cite{yuan2025debug,ding2026octobench}}
& T & E & I & T & E & E & -- & -- \\

\mbox{BashArena / ClawSafety~\cite{kaufman2025basharena,wei2026clawsafety}}
& E & E & I & T & T & T & E & -- \\

LiveSQLBench~\cite{bird2024livesqlbench}
& T & T & -- & I & -- & -- & -- & E \\

WildClawBench~\cite{ding2026wildclawbench}
& E & E & T & E & T & T & I & -- \\

ClawForge~\cite{lai2026clawforge}
& E & E & T & E & T & I & -- & -- \\

\bottomrule
\end{tabularx}
\end{table*}

The matrix shows broad observability of command formulation and feedback interpretation, especially in terminal-native and CLI-centered benchmarks. Runtime management is explicit in SetupBench~\cite{arora2025setupbench} and trace-visible in Terminal-Bench~\cite{merrill2026terminal}, TerminalWorld~\cite{chu2026terminalworld}, CLI-Gym~\cite{lin2026cli}, WildClawBench~\cite{ding2026wildclawbench}, and ClawForge~\cite{lai2026clawforge}. Long-horizon and deployment-like settings expose state and context tracking, often indirectly through final outcomes. Debugging and process benchmarks provide stronger observability of verification and recovery~\cite{yuan2025debug,ding2026octobench}, whereas repository-level evaluations rarely isolate them. Governance is explicit mainly in BashArena~\cite{kaufman2025basharena} and ClawSafety~\cite{wei2026clawsafety}; freshness remains a separate validity property.

\subsection{Remaining Evaluation Gaps}
\label{subsec:evaluation_gaps}

Four structural gaps remain. \textbf{Terminal-native workflow coverage is narrow.} Most benchmarks remain repository- or coding-centric, while alternatives are fragmented across operational domains. TerminalWorld broadens general terminal work~\cite{chu2026terminalworld}; ML-DevBench and MLE-bench target iterative model development~\cite{padigela2025ml,chan2025mle}; ELTBench, DAComp, and DSAgentBench cover data pipelines and end-to-end data-science workflows~\cite{jin2025elt,lei2025dacomp,rahman2026dsagentbench}; and ITBench and AIOpsLab introduce operational state~\cite{jha2025itbench,chen2025aiopslab}. Scientific experimentation appears in ExpBench, Curie, and ScienceBoard~\cite{kon2025exp,kon2025curie,sun2025scienceboard}, while security evaluation covers CTF, penetration testing, and inference optimization~\cite{lee2026ctfusion,al2026agents,liu2025pacebench,nangia2026iso}. End-to-end CLI tool generation also remains under-evaluated~\cite{hu2026evaluating}.

\textbf{Process and trace standards are immature.} Benchmarks increasingly record trajectories, but no common schema covers commands, observations, failures, retries, state changes, and human interventions. Trajectory analyses demonstrate the value of trace inspection~\cite{bouzenia2025understanding,ceka2025understanding}; ProcBench adds step-level assessment~\cite{he2026procbench}; and IDE-Bench extends evaluation to IDE settings~\cite{mateega2026ide}. Studies of failed agentic pull requests and CLI trajectories further motivate shared failure taxonomies~\cite{ehsani2026ai,zhao2026failureprocess}. Interactive trajectory debugging remains disconnected from benchmark protocols~\cite{hutter2026agentstepper}.

\textbf{Freshness mechanisms remain peripheral.} SWE-rebench reports evidence consistent with contamination-related inflation on static tasks~\cite{badertdinov2026swerebench}. Mutation and detection methods are emerging~\cite{garg2025saving,song2026cross,chai2026benchmarks}, but rarely enter standard evaluation pipelines, while temporal-consistency mechanisms remain experimental~\cite{sun2026atime}.

\textbf{Safety and governance remain separated from task success.} Existing benchmarks isolate privileged terminal actions~\cite{kaufman2025basharena}; risky code generation and execution~\cite{guo2024redcode,chen2026securevibebenchbenchmarkingsecurevibe,dawson2025airtbench,ba2026ciber}; productivity and computer-use agents~\cite{li2026clawsbench,feng2026agenthazard,kuntz2026harm,chen2026lps}; and attacks on long-running or sandboxed agents~\cite{marchand2026quantifying,ye2026claw,song2026anchor}. UnderSpecBench extends earlier evidence of governance blind spots under benign instructions by measuring wrong-target and over-scope actions in underspecified DevOps tasks, while Boundary-Bench evaluates coding agents under progressively hardened execution policies~\cite{ding2026blind,ji2026codingagentsguessingmeasuring,davidovich2026permissiondeniedpolicygradedevaluation}. These protocols make authorization and containment observable, but governance evidence remains distributed across separate task-success and policy-focused settings.

Current evaluation suites distribute these requirements across separate protocols rather than jointly integrating workflow realism, environment setup, process quality, trace visibility, long-horizon persistence, freshness, safety, and governance~\cite{arora2025setupbench,he2026procbench,kuang2025process,badertdinov2026swerebench,kaufman2025basharena,wei2026clawsafety}. Evaluation results therefore characterize a coupled configuration defined by the model, benchmark, harness, runtime, sandbox, observation format, and execution budget, while trace-release policy determines how much process evidence remains available for analysis.

Architecture shapes what can be executed and recorded, acquisition determines which recorded behavior can be learned, and evaluation determines which behaviors become evidence. Section~\ref{sec:cross_cutting_insights} uses bounded fixed-condition diagnostics to illustrate two consequences: benchmark-dependent process exposure and limits of component attribution.

%% file: sections/6_empirical_analysis_insights.tex
\section{Framework-Guided Diagnostics of Process Observability and Attribution Limits}
\label{sec:cross_cutting_insights}

The preceding sections connect terminal competence to system architecture, acquisition data, and evaluation protocols. This section makes two implications of that synthesis concrete through bounded fixed-condition diagnostics. First, under a fixed agent configuration, we examine which process signals different benchmark families expose. Second, on matched repository-repair tasks, we examine how outcomes vary across systems and what these differences reveal about component attribution. Trace cases complement the aggregate results with trajectory-level mechanisms.

For the benchmark-exposure diagnostic, we fix mini-SWE-agent~\cite{yang2024swe} with DeepSeek-V4-Flash~\cite{xu2026deepseek} while varying four benchmark families. The matched diagnostic separately evaluates mini-SWE-agent~\cite{yang2024swe}, SWE-agent~\cite{yang2024swe}, and OpenHands~\cite{wang2025openhands} on identical task identifiers within Claw-SWE-Bench Lite~\cite{zheng2026claw} and SWE-bench Lite~\cite{jimenez2024swe}, using DeepSeek-V4-Flash and DeepSeek-V4-Pro~\cite{xu2026deepseek}. Each task has one formal run per condition, and each cell represents a complete model, interface, harness, and runtime configuration.

\subsection{Diagnostic Design and Measures}
\label{subsec:empirical_protocol}

Both diagnostics retain benchmark-native outcomes and trace-level process evidence. Seven trace-derived indicators, denoted P1--P7, capture selected observable aspects associated with the competence dimensions introduced in Section~\ref{sec:background_terminal_agents}. The identifiers provide concise cross-reference and are paired with semantic names throughout the analysis.

Four indicators are deterministic. The \emph{rule-matched invocation-failure rate} (P1) divides failed commands whose stderr matches shell syntax, command-not-found, non-executable, path, permission, argument, or malformed-tool patterns by all normalized command events. The \emph{environment-exit rate} (P3) divides non-zero exits on dependency, runtime, service, configuration, or environment-management commands by all detected environment-management commands. The \emph{final-window verification rate} (P5) is the fraction of tasks with a detected verification action in the final five actions or final 20\% of the trace, whichever window is larger. The \emph{governance-review-trigger rate} (P7) divides command events matching irreversible, overprivileged, secret-handling, host or sandbox, or external-side-effect patterns by all normalized command events.

Three auxiliary indicators use rule-constrained LLM judgments. Rule-based extractors first identify candidate episodes concerning feedback use, state errors, or recovery, and DeepSeek-V4-Flash~\cite{xu2026deepseek} evaluates each target within a bounded evidence window. P2 is the helpful-use rate among eligible episodes with usable feedback, P4 the consequential-error rate among episodes with decidable state evidence, and P6 the successful task-relevant recovery rate among episodes containing a recovery attempt. Uncertain or low-confidence labels are excluded, and task-level rates are macro-averaged over eligible evidence.

All formal runs use fixed decoding settings, task identifiers, and benchmark-native evaluators. Harness-specific prompting, context management, action interfaces, and orchestration remain part of each evaluated system. Complete model settings, execution budgets, extractor definitions, and auxiliary-judge procedures are reported in the Supplementary Material.

The benchmark-exposure profile contains 241 Terminal-Bench 2.1 tasks~\cite{merrill2026terminal}, 93 SetupBench tasks~\cite{arora2025setupbench}, 21 LongCLI-Bench tasks~\cite{feng2026longcli}, and 640 BashArena tasks~\cite{kaufman2025basharena}. The matched system comparison contains all 80 official Claw-SWE-Bench Lite tasks and 300 SWE-bench Lite tasks.

\subsection{Benchmark-Exposure Diagnostic}
\label{subsec:fixed_harness_capability_profile}

Table~\ref{tab:fixed_harness_capability_profile} reports benchmark-native outcomes and the seven trace-derived process indicators. Because the benchmarks use different task distributions and evaluators, outcomes are interpreted within each benchmark rather than as a shared competence scale. P1, P3, P5, and P7 are deterministic, whereas P2, P4, and P6 are judge-assisted task-level macro averages over eligible semantic evidence.

\begin{table*}[!t]
\caption{Benchmark-exposure profile under mini-SWE-agent with DeepSeek-V4-Flash. Outcome is the official benchmark score. P1--P7 are bounded trace-derived process indicators. P1 and P3 record execution-friction signals; P2, P4, and P6 are auxiliary judge-assisted task-level rates; P5 records a detected final-window check; and P7 flags actions requiring contextual governance review.}
\label{tab:fixed_harness_capability_profile}
\centering
\scriptsize
\setlength{\tabcolsep}{2.4pt}
\renewcommand{\arraystretch}{1.07}
\begin{tabularx}{\textwidth}{
@{}
>{\raggedright\arraybackslash}p{2.85cm}
>{\centering\arraybackslash}p{0.65cm}
>{\centering\arraybackslash}p{0.9cm}
*{7}{>{\centering\arraybackslash}X}
@{}}
\toprule
Benchmark
& Tasks
& Outcome
& \makecell{P1\\Rule inv. fail.}
& \makecell{P2\\Feedback}
& \makecell{P3\\Env. exit}
& \makecell{P4\\State err.}
& \makecell{P5\\Final verify}
& \makecell{P6\\Recovery}
& \makecell{P7\\Gov. trigger}
\\
\midrule
Terminal-Bench 2.1~\cite{merrill2026terminal}
& 241 & 52.6\% & 0.0\% & 82.1\% & 12.5\% & 14.7\% & 29.5\% & 79.9\% & 0.5\% \\
SetupBench~\cite{arora2025setupbench}
& 93 & 59.1\% & 0.0\% & 80.2\% & 9.0\% & 19.5\% & 36.6\% & 85.5\% & 1.4\% \\
LongCLI-Bench~\cite{feng2026longcli}
& 21 & 23.8\% & 0.0\% & 76.4\% & 6.8\% & 7.7\% & 23.8\% & 67.8\% & 4.5\% \\
BashArena~\cite{kaufman2025basharena}
& 640 & 41.8\% & 0.0\% & 76.3\% & 13.7\% & 23.3\% & 66.2\% & 67.0\% & 3.0\% \\
\bottomrule
\end{tabularx}
\end{table*}

Three findings characterize the observed profile. \textbf{Rule-matched invocation failures are rare under the fixed configuration.} P1 rounds to 0.0\% across all four benchmarks, indicating that few command failures match the specified shell syntax, command-not-found, non-executable, path, permission, argument, or malformed-tool patterns. Broader action-formulation problems instead appear through semantic or state-dependent behavior not captured by this narrow signal.

\textbf{Local feedback use and end-to-end completion expose different aspects of behavior.} The auxiliary helpful-feedback indicator (P2) ranges from 76.3\% to 82.1\%, showing that many eligible episodes contain a locally useful response to visible feedback. Benchmark-native outcomes additionally depend on runtime management, persistent state tracking, recovery, and verification. Detected final-window verification ranges from 23.8\% to 36.6\% for Terminal-Bench, SetupBench, and LongCLI-Bench, compared with 66.2\% for BashArena, showing substantial differences in late-trace inspection across benchmark conditions.

\textbf{Benchmark conditions foreground different process limitations.} Under the fixed configuration, SetupBench foregrounds environment bootstrap, LongCLI-Bench combines low outcome with long-horizon interaction demands, and BashArena increases the visibility of state errors and governance-relevant actions. The auxiliary indicators show broadly high accepted-label coverage, although LongCLI-Bench has both the smallest task set and lower semantic coverage than the other benchmarks. Its P2, P4, and P6 estimates are therefore interpreted directionally.

\subsection{Matched Complete-System Diagnostic}
\label{subsec:controlled_harness_effects}

The second diagnostic focuses on outcome differences across systems rather than extending the P1--P7 profile. Table~\ref{tab:controlled_harness_effects} reports matched execution snapshots on identical task identifiers within each benchmark.

SWE-agent has the highest resolved rate in all four benchmark--model blocks. Its largest gap to the lowest-performing system is 21.25 percentage points on Claw-SWE-Bench Lite, compared with at most 7.00 points on SWE-bench Lite. The relative ordering of OpenHands and mini-SWE-agent also reverses across benchmarks: OpenHands is higher on Claw-SWE-Bench Lite but lower on SWE-bench Lite. Benchmark choice therefore changes the observed separation among systems even though SWE-agent remains highest under all reported conditions.

\begin{table}[!t]
\caption{Matched single-run results by benchmark and system. F/P denote DeepSeek-V4-Flash/Pro; entries report resolved rate, Pro-minus-Flash difference, and mean task time in seconds.}
\label{tab:controlled_harness_effects}
\centering
\scriptsize
\setlength{\tabcolsep}{3.4pt}
\renewcommand{\arraystretch}{1.05}
\begin{tabular*}{\columnwidth}{@{\extracolsep{\fill}}lcccc@{}}
\toprule
System & F & P & $\Delta$ & Time F/P \\
\midrule
\multicolumn{5}{c}{\emph{Claw-SWE-Bench Lite}} \\
\cmidrule(lr){1-5}
mini-SWE-agent & 58.75\% & 56.25\% & $-2.50$ & 476/476 \\
SWE-agent~\cite{yang2024swe} & \textbf{76.25\%} & \textbf{77.50\%} & $+1.25$ & 675/712 \\
OpenHands~\cite{wang2025openhands} & 68.75\% & 67.50\% & $-1.25$ & 600/653 \\
\midrule
\multicolumn{5}{c}{\emph{SWE-bench Lite}} \\
\cmidrule(lr){1-5}
mini-SWE-agent & 55.67\% & 55.67\% & $0.00$ & 294/389 \\
SWE-agent~\cite{yang2024swe} & \textbf{58.67\%} & \textbf{58.33\%} & $-0.33$ & 382/525 \\
OpenHands~\cite{wang2025openhands} & 51.67\% & 51.67\% & $0.00$ & 501/504 \\
\bottomrule
\end{tabular*}
\end{table}

Within a fixed system, the absolute Pro-minus-Flash difference never exceeds 2.50 points. Exact task-paired McNemar tests detect no directional model-variant difference after Holm adjustment, and all paired bootstrap intervals include zero. Under these conditions, the two model variants produce similar resolved rates across the evaluated systems. Pro has higher observed mean wall-clock time in five of six system--benchmark pairs; under mini-SWE-agent, it also uses more mean input tokens on both benchmarks without an outcome gain.

Task-paired cross-system tests further characterize these snapshots. Cochran's $Q$ rejects equal resolved rates among the three systems in every benchmark--model block ($Q=10.57$ to $16.00$, $p<0.006$). After Holm correction across 12 pairwise comparisons, SWE-agent differs from mini-SWE-agent on Claw-SWE-Bench Lite for both Flash ($p=0.023$) and Pro ($p=0.002$), and from OpenHands on SWE-bench Lite for both Flash ($p=0.001$) and Pro ($p=0.001$). The other corrected pairwise contrasts are not significant. These results characterize task-level differences within the recorded execution snapshots.

\subsection{Illustrative Trace Cases}
\label{subsec:trace_metrics_failure_cases}

Aggregate rates alone do not show how these differences arise within trajectories. Two trace cases illustrate how benchmark demands and system behavior become visible at the execution level.

\begin{evidencebox}{Benchmark-exposure case: \texttt{magsac-install}}
\textbf{Condition:} Terminal-Bench 2.1 with mini-SWE-agent and DeepSeek-V4-Flash.
\textbf{Outcome:} unresolved; the evaluator reports \texttt{No module named 'pymagsac'}.
\textbf{Trace evidence:} repeated dependency and build attempts yield a 38.5\% environment-command non-zero-exit rate. The trace ends before an evaluator-relevant import or installation-source check.
\textbf{Interpretation:} local feedback leaves the installation loop unclosed, linking runtime management, verification, and recovery across the trajectory.
\end{evidencebox}

Together, the aggregate results and trace cases show that benchmark conditions change which process demands become visible, while system configuration changes realized trajectories and outcomes on matched tasks. These diagnostics connect the survey framework to observable execution behavior and motivate evaluation that combines task outcomes with process evidence and explicit system conditions.

\begin{evidencebox}{Matched case: \texttt{rubocop\_\_rubocop-13560}}
\textbf{Condition:} Claw-SWE-Bench Lite with DeepSeek-V4-Flash.
\textbf{Outcome:} SWE-agent resolves the task; mini-SWE-agent and OpenHands do not.
\textbf{Trace evidence:} the issue requires ordinary lowercase \texttt{'nul'} arguments to remain valid. Mini-SWE-agent changes the implementation and an existing test despite this constraint. OpenHands exempts all method-call arguments, broadening the requested behavior. SWE-agent changes production logic while preserving the required distinction and passes the official evaluator.
\textbf{Interpretation:} the task exposes different patch scopes and outcomes across systems.
\end{evidencebox}

\FloatBarrier

%% file: sections/7_challenges_future.tex
\section{Challenges and Future Directions}
\label{sec:challenges_future}

The literature synthesis identifies four connected gaps: domain concentration limits evidence for transferable competence, fragmented process evaluation restricts behavioral interpretation, runtime governance remains weakly integrated with task evaluation, and system-level comparisons complicate component attribution. These gaps recur across the architecture, acquisition, and evaluation evidence reviewed in Sections~\ref{sec:terminal_agent_system_design_architectures}--\ref{sec:benchmarks_metrics_evaluation} and are further illustrated by the process-observability and attribution diagnostics in Section~\ref{sec:cross_cutting_insights}, motivating four research priorities.

\subsection{Cross-Domain Terminal Competence}
\label{subsec:cross_domain_terminal_competence}

Training and evaluation remain concentrated in software engineering~\cite{jimenez2024swe,yang2024swe,pan2024training}. Repositories provide structured executable verification, but cannot determine whether terminal behavior reflects transferable command and interaction competence or task-specific heuristics. Operations, data engineering, scientific workflows, cloud management, and cybersecurity introduce distinct runtime and governance demands~\cite{ardebili2025kubeintellect,siva2026kraig,twabi2026netagentbench,jin2025elt,kon2025curie,chen2025aiopslab,kaufman2025basharena}, while TUA-Bench extends evaluation to routine digital activities and scientific and engineering workflows~\cite{chen2026tuabenchbenchmarkgeneralpurposeterminaluse}. Cross-domain research should distinguish reusable substrate-level competence from domain-specific workflows and compare transfer across competence dimensions to identify which capabilities remain workload-specific.

\subsection{Fresh, Replayable, Process-Level Evaluation}
\label{subsec:fresh_replayable_process_eval}

Final outcomes do not reveal diagnostic quality, state preservation, recovery, or unsafe intermediate behavior~\cite{ding2026octobench,he2026procbench,kuang2025process}. Live or regenerated tasks improve temporal validity, while replayable traces preserve commands, observations, state changes, interventions, and verifier calls~\cite{badertdinov2026swerebench,bouzenia2025understanding,ceka2025understanding}. Combining them would connect fresh task distributions with process evidence, distinguishing successful completion from the behavior that produces it. This also links evaluation to acquisition in Section~\ref{sec:terminal_competence_acquisition}, where retained failures, state transitions, and recovery trajectories determine what can be learned and inspected.

\subsection{Runtime Governability and Safety}
\label{subsec:runtime_governability_safety}

Terminal agents can modify filesystems, install packages, launch processes, access networks, and operate around credentials. Permission gates, sandboxes, approvals, and rollback constrain these effects, but governance remains distributed across separate safety and policy-focused protocols~\cite{andriushchenko2025agentharm,kaufman2025basharena,wei2026clawsafety,lee2026ctfusion,ji2026codingagentsguessingmeasuring,davidovich2026permissiondeniedpolicygradedevaluation}. Future evaluation should integrate the architectural controls identified in Section~\ref{sec:terminal_agent_system_design_architectures}, including authorization, containment, reversibility, destructive-action prevention, and external side-effect control. Task success and governance should be assessed jointly so that effective execution reflects the constraints under which actions are taken.

\subsection{Controlled Model and Harness Attribution}
\label{subsec:controlled_model_harness_attribution}

Measured performance can reflect the model, interface, runtime, context policy, retry budget, or verifier access~\cite{yang2024swe,wang2025openhands,lee2026meta,lou2026autoharness}. Controlled studies further show that harness and LM--CLI choices can alter quality, efficiency, and failure behavior~\cite{vats2026scaffoldeffectcodingagents,chi2026matchingmattersfairqualityefficiency}. Separating model capability from surrounding system support therefore remains a central problem. The matched diagnostic in Section~\ref{sec:cross_cutting_insights} illustrates that observed system differences can also vary across benchmarks. Factorial designs or portable protocols that vary the model, interface, harness, or runtime while preserving task identity and execution conditions~\cite{thangarajah2026dcas} can better isolate which gains transfer across configurations and which arise from specific model--harness interactions.

%% file: sections/8_conclusion.tex
\section{Conclusion}
\label{sec:conclusion}

We organize the study of terminal agents around terminal-mediated execution and its state-changing action--observation loop, using a seven-dimensional competence profile to connect system architecture, competence acquisition, and evaluation.

Three conclusions emerge. First, terminal-agent behavior is jointly shaped by the model, interface, harness, runtime, and environment, making outer-loop design a performance-shaping system component. Second, executable trajectories ground learning in feedback, verification, and recovery, yet current acquisition remains concentrated on successful repository workflows and underrepresents environment management, persistent state, recovery, and governance. Third, prevailing evaluations emphasize final outcomes and expose process quality unevenly, while measured performance depends on both benchmark conditions and system configuration. The bounded diagnostics further illustrate benchmark-dependent process exposure and limits of component attribution.

Progress therefore requires cross-domain acquisition and evaluation, fresh and replayable process evidence, governable runtimes, and controlled model--harness attribution. Evaluation should combine explicit system and runtime conditions with task outcomes, process evidence, and trace provenance.

%% file: sections/appendix_methodology.tex
\section{Review Protocol and Corpus Construction}
\label{app:review_protocol}

This supplementary document reports the review protocol, extended taxonomy and coding materials, empirical methods and results, and representative trace cases. We conducted a \textbf{structured narrative review with an evidence-stratified coded corpus}. The review combines explicit workload-level scope rules, entry-level analytical coding, and passage-level evidence calibration to support a traceable synthesis of the reconciled corpus.

\subsection{Corpus Scope and Composition}

The review covers work released between 2022 and August 2026 on terminal agents, terminal-mediated architectures, executable training environments, terminal-native or repository-based evaluation, process-level agent assessment, runtime governance, and adjacent agent paradigms used for boundary comparison. The current manuscript cites \textbf{191 distinct entries}: \textbf{186 research entries} in the review corpus, four deployment-facing tools (Claude Code, Codex CLI, Aider, and Gemini CLI), one CLI-packaged assistant boundary comparator. The latter six entries remain outside research statistics. Candidate versions are consolidated at the work level so that the corpus counts a research contribution rather than each of its releases.

Table~\ref{tab:corpus_summary} summarizes corpus status, release year, and analytical use. Analytical-use counts are non-exclusive because one work may support several parts of the survey. This view separates entry-level corpus composition from subset-specific architecture, acquisition, and evaluation coding and from passage-level evidence calibration.

\begin{table*}[!t]
\caption{Composition of the current review corpus. Release-year counts cover the 186 research entries; analytical-use counts are non-exclusive.}
\label{tab:corpus_summary}
\centering
\footnotesize
\setlength{\tabcolsep}{5pt}
\renewcommand{\arraystretch}{1.08}
\begin{tabularx}{\textwidth}{@{}>{\raggedright\arraybackslash}p{3.1cm}X>{\raggedleft\arraybackslash}p{0.9cm}@{}}
\toprule
\textbf{Category} & \textbf{Description} & \textbf{Count} \\
\midrule

\multicolumn{3}{@{}l}{\textbf{Corpus status}} \\
Research corpus & Research works retained for the survey synthesis. & 186 \\
Engineering practice, boundary, and disclosure references & Five engineering-practice or boundary entries and one disclosure citation. & 6 \\
Total cited set & Distinct cited entries after work-level version consolidation. & 192 \\

\addlinespace[2pt]
\midrule
\multicolumn{3}{@{}l}{\textbf{Release year of research entries}} \\
2022--2023 & Earliest retained foundations for tool use and interactive execution. & 4 \\
2024 & Expansion of executable agent and repository-repair research. & 14 \\
2025 & Broadening benchmark, system, and deployment evidence. & 56 \\
2026 through August & Recent architecture, acquisition, evaluation, and cross-domain work. & 112 \\

\addlinespace[2pt]
\midrule
\multicolumn{3}{@{}l}{\textbf{Analytical use of research entries, non-exclusive}} \\
Scope and characterization & Supports the execution-substrate boundary and competence profile. & 19 \\
System architecture & Supports layers, families, control, memory, and runtime analysis. & 44 \\
Competence acquisition & Supports environments, trajectories, supervision, and adaptation analysis. & 41 \\
Evaluation & Supports benchmark families, metrics, validity, and competence observability. & 99 \\
Empirical setting & Supports the fixed-condition diagnostic design and its task settings. & 9 \\
Challenges and implications & Supports domain transfer, process evidence, governance, and attribution. & 28 \\

\bottomrule
\end{tabularx}
\end{table*}

\subsection{Search, Screening, and Work-Level Reconciliation}

We assembled the corpus through iterative keyword search, venue-oriented search, and backward and forward reference chaining, with the final update performed in August 2026. Sources included arXiv, OpenReview, DBLP, IEEE Xplore, the ACM Digital Library, the ACL Anthology, major machine-learning and software-engineering venues, systems-oriented venues, benchmark repositories, and project pages for deployment-facing terminal-agent tools. Candidate versions were consolidated at the work level, with the most complete version retained and earlier versions linked to the same contribution rather than counted separately.

Table~\ref{sup:review_workflow} summarizes the construction sequence and its analytical role. The sequence also provides the protocol used for subsequent corpus updates.

\begin{table*}[!t]
\caption{Corpus-construction workflow. Each stage produces the input required by the next stage and constrains how evidence enters the synthesis.}
\label{sup:review_workflow}
\centering
\footnotesize
\setlength{\tabcolsep}{4pt}
\renewcommand{\arraystretch}{1.00}
\begin{tabularx}{\textwidth}{@{}p{2.15cm}p{4.1cm}p{4.0cm}X@{}}
\toprule
Stage & Procedure & Recorded decision & Role in synthesis \\
\midrule
Scope freeze & Fix the 2022 to August 2026 interval, terminal-mediated research object, adjacent comparators, and workload-level boundary tests & Date range, source classes, and inclusion boundary & Prevents surface-level CLI mentions from entering as core evidence \\
Retrieval & Apply keyword families, venue-oriented search, and backward and forward reference chaining & Source, execution date, query form when retained, and work identity & Combines named-field retrieval with architecture, acquisition, evaluation, and governance routes \\
Work reconciliation & Group versions of the same contribution and retain the most complete research account & Preferred version and version relationship & Avoids double counting preprint, conference, and journal releases \\
Screening & Apply the five questions below and assign a corpus disposition & Research, engineering-practice, boundary, or exclusion disposition & Separates progress-bearing terminal evidence from surface-level or adjacent evidence \\
Analytical coding & Assign manuscript roles to every retained entry and apply architecture, acquisition, or evaluation fields to the relevant subsets & Entry-level roles and subset-specific analytical codes & Connects retained evidence to the architecture, acquisition, and evaluation synthesis \\
Claim calibration & Assess convergence, directness, domain breadth, and evidence maturity at the passage level & Bounded claim strength and scope & Keeps synthesis claims proportional to the available evidence \\
Update check & Reapply retrieval, reconcile new versions, repeat screening, and revise affected synthesis passages & Added, consolidated, retained, or removed status & Keeps later corpus updates aligned with the same decision sequence \\
\bottomrule
\end{tabularx}
\end{table*}

For subsequent updates, we specify seven Boolean query families in Table~\ref{sup:query_families}. Source interfaces may require field-name or quotation-mark changes, while the Boolean concepts and date limits remain fixed. Each new source-specific execution records its date, query form, and hit count before work-level reconciliation.

\begin{table*}[!t]
\caption{Boolean query families used for subsequent corpus updates. Each source-specific execution is logged with its date and hit count.}
\label{sup:query_families}
\centering
\scriptsize
\setlength{\tabcolsep}{3.2pt}
\begin{tabularx}{\textwidth}{@{}p{0.55cm}p{2.6cm}X@{}}
\toprule
ID & Purpose & Boolean template \\
\midrule
Q1 & Named field & \texttt{("terminal agent" OR "terminal agents" OR "CLI agent" OR "command-line agent")} \\
Q2 & Execution loop & \texttt{("LLM agent" OR "language model agent") AND (terminal OR shell OR CLI OR "command line") AND (execution OR runtime OR feedback)} \\
Q3 & SWE and harnesses & \texttt{("software engineering agent" OR "coding agent") AND (terminal OR shell OR command) AND (benchmark OR harness OR environment)} \\
Q4 & Acquisition & \texttt{("agent trajectory" OR "executable trajectory" OR "environment interaction") AND (training OR post-training OR "reinforcement learning") AND (terminal OR CLI OR repository)} \\
Q5 & Process evaluation & \texttt{("terminal benchmark" OR "CLI benchmark" OR "agent benchmark") AND (trace OR process OR recovery OR verification OR state)} \\
Q6 & Governance & \texttt{("LLM agent" OR "coding agent" OR "computer-use agent") AND (shell OR terminal OR "code execution") AND (sandbox OR permission OR safety OR governance)} \\
Q7 & Cross-domain use & \texttt{("LLM agent" OR "AI agent") AND (terminal OR CLI OR "command execution") AND (AIOps OR CloudOps OR DataOps OR cybersecurity OR "scientific workflow")} \\
\bottomrule
\end{tabularx}
\end{table*}

Each candidate work was screened using five questions:
\begin{enumerate}[leftmargin=*,itemsep=1pt,topsep=2pt]
    \item Does the agent execute terminal commands, operate CLI tools, or interact with a terminal-mediated environment?
    \item Does stdout, stderr, logs, diffs, return codes, or execution feedback materially shape subsequent actions?
    \item Does the system produce real or simulated environment state changes through execution?
    \item Does the work provide executable verification, trajectory data, a benchmark, an acquisition pipeline, or a runtime architecture relevant to terminal-mediated execution or terminal-agent systems?
    \item What claim scope does the work support: core terminal-agent evidence, terminal-hybrid evidence, executable SWE-adjacent evidence, boundary comparison, or background framing?
\end{enumerate}

Based on these questions, each retained entry received a corpus status and one or more analytical roles. Research entries support the architecture, acquisition, evaluation, empirical, or implication synthesis; engineering-practice and boundary references illustrate deployed patterns or delimit adjacent systems. The reported corpus counts begin after work-level reconciliation because the iterative initial retrieval did not preserve a complete query-level candidate count before deduplication.

\subsection{Corpus Reconciliation and Update Protocol}

The retained corpus can be inspected through the screening questions, work-level reconciliation rules, analytical roles, and level-specific fields in Table~\ref{sup:coding_schema}. Subsequent updates additionally use the Boolean query families in Table~\ref{sup:query_families}. Retrieval results are reconciled at the work level before assignment of corpus disposition and analytical roles. Applying these rules yields the 186-entry research corpus summarized in Table~\ref{tab:corpus_summary}; the six non-corpus references remain outside research-entry statistics.

During an update, a work is added when it satisfies the screening questions and contributes evidence to at least one analytical role. A new version replaces an earlier version when it provides the more complete account of the same contribution. Versions representing the same contribution are consolidated rather than counted separately. A retained work is removed from the review corpus when it no longer supports a synthesis passage after manuscript revision. These rules keep corpus composition, analytical coding, and manuscript claims synchronized.

\subsection{Coding Dimensions and Evidence Calibration}

Entry-level coding covers bibliographic metadata, research or engineering-practice status, and manuscript analytical roles. These fields support the corpus counts, release-year distribution, and non-exclusive analytical-use counts in Table~\ref{tab:corpus_summary}. Additional conceptual fields are applied only to evidence subsets for which they are relevant, using the paper as the default unit and the system, benchmark, dataset, or pipeline when one paper introduces several distinct artifacts.

For the main paper's architecture analysis, relevant entries were coded by interface granularity, operational coupling, autonomy regime, recovery and control style, and planning or control strategy. For the acquisition analysis, entries were coded by data-source type, trajectory source, supervision signal, filtering or relabeling method, failure-data treatment, and capability target. For the evaluation synthesis and diagnostic study, entries were coded by task regime, harness assumptions, verification type, trace visibility, artifact availability, and coverage of the seven terminal-competence dimensions.

We calibrate evidence at the synthesis-passage level rather than assigning a single maturity label to an entire paper. Converging evidence across multiple benchmarks, systems, or domains supports stronger synthesis claims; direct but limited empirical evidence supports bounded claims; single-study or indirect evidence supports provisional claims; and deployment practice or boundary examples support illustration. Publication status and artifact availability remain descriptors of the relevant evidence subset rather than substitutes for claim-level calibration.

\begin{table*}[!t]
\caption{Level-specific coding schema used in corpus construction and synthesis. Entry-level fields cover every retained reference; specialized fields apply to relevant evidence subsets or synthesis passages.}
\label{sup:coding_schema}
\centering
\footnotesize
\setlength{\tabcolsep}{4pt}
\renewcommand{\arraystretch}{1.08}
\begin{tabularx}{\textwidth}{@{}p{2.5cm}p{5.0cm}p{2.25cm}X@{}}
\toprule
Field group & Coded fields & Unit & Analytical use \\
\midrule
Corpus identity & Bibliographic metadata, corpus status, and unique work identity & Work & Supports corpus identity, release-year counts, and work-level deduplication \\
Analytical role & Scope, architecture, acquisition, evaluation, empirical setting, or implications & Work, non-exclusive & Connects retained entries to the survey passages they support \\
Architecture & Interface granularity, operational coupling, autonomy regime, recovery and control style, planning strategy & System or artifact & Supports the layered architecture and recurring-pattern synthesis \\
Acquisition & Data source, trajectory source, supervision signal, filtering or relabeling, failure-data treatment, capability target & Dataset, environment, or pipeline & Supports comparison of executable data, learning, and adaptation routes \\
Evaluation & Task regime, harness assumptions, verification type, trace visibility, artifact availability, dimension coverage & Benchmark or protocol & Supports benchmark-family and competence-observability comparisons \\
Evidence calibration & Convergence, directness, domain breadth, and maturity of evidence for the passage claim & Synthesis passage & Bounds the strength and generality of each synthesized claim \\
\bottomrule
\end{tabularx}
\end{table*}

\subsection{Terminology and System-Layer Roles}

The working terminology used for the survey's workload-level boundary tests is organized as follows. Each term is paired with its technical meaning and its analytical role.
\compactentry{Terminal.} A textual input/output access layer or its emulation. It is an established field label and a common access path to command execution, not itself the source of process state or exit semantics.
\compactentry{Shell.} A command interpreter that parses commands, expands syntax, and starts programs. It is a common but non-required mediator between an agent interface and executable programs.
\compactentry{CLI tool.} A program whose operations are invoked through textual arguments or commands. It is an action target; surface-level CLI exposure alone does not place a workload in scope.
\compactentry{Command-execution runtime.} The filesystem, processes, dependencies, permissions, resources, and operating-system state in which commands take effect. It is the state-changing execution substrate that grounds actions, observations, and later verification.
\compactentry{Harness.} The model-facing layer that exposes actions, formats observations, manages context, invokes the runtime, and applies execution policy. It allocates system responsibilities and changes what the model can observe or do.
\compactentry{Terminal agent.} A system whose dominant progress-bearing loop depends on command execution, textual feedback, and stateful environment interaction. This is the survey's workload-level working characterization, applied through the three boundary tests.

\subsection{Construction and Boundaries of the Seven-Dimension Profile}

The seven dimensions synthesize recurring functional responsibilities and their cross-dimensional dependencies. We first collected responsibilities and failure descriptions from architecture, acquisition, benchmark, and process-evaluation studies. We then grouped them by the object acted upon, the evidence needed for the next decision, and the response required when execution diverged from the task. Candidate groupings were separated when they required distinct system mechanisms or observable evidence. The resulting construction logic and representative trace signals are summarized below.
\compactentry{Command and action formulation.} Select commands, arguments, edits, and executable sequences. The separation rule concerns the action submitted before interpreting its consequence. Representative failures are malformed invocation, wrong target, and semantically unsuitable action. Commands and tool calls reveal form, while intent and semantic suitability may remain latent.
\compactentry{Feedback and artifact interpretation.} Extract decision-relevant evidence from outputs and artifacts. The separation rule concerns how observed evidence informs the next decision. Representative failures are ignored error, superficial reaction, and misread test or diff. Action changes after feedback are visible, while usefulness requires contextual judgment.
\compactentry{Runtime and environment management.} Construct and maintain dependencies, services, processes, and execution conditions. The separation rule concerns the environment required for actions to run rather than the remembered task state. Representative failures are dependency conflict, service failure, and unstable configuration. Exit codes and setup actions expose friction, but successful exits do not prove a correct runtime.
\compactentry{State, task, and context tracking.} Maintain facts about workspace, task, history, and unresolved goals. The separation rule concerns the agent's working account of persistent state across steps. Representative failures are stale path, forgotten constraint, repeated loop, and inconsistent goal. Contradictions and repeated errors can be traced, but complete internal state is not observable.
\compactentry{Progress verification.} Design checks that establish intermediate validity or completion. The separation rule concerns evidence that a claim or artifact satisfies a condition. Representative failures are missing check, inadequate oracle, and unchecked requirement. Tests and inspections expose verification activity, while timing or presence does not establish adequacy.
\compactentry{Recovery and adaptation.} Diagnose divergence, revise strategy, retry, work around, or roll back. This dimension begins after observed failure or violated expectation and targets renewed progress. Representative failures are repeated ineffective retry, irrelevant workaround, and failed rollback. Failure-follow-up windows expose attempts, while causal relevance and success require context.
\compactentry{Governance and side-effect control.} Respect authorization, containment, reversibility, credentials, and resource limits. The separation rule concerns whether execution remains within policy independent of command success. Representative failures are unauthorized deletion, secret exposure, sandbox escape, and uncontrolled side effect. Sensitive events trigger review, while task authorization is needed to determine a violation.

We distinguish runtime and environment management from state, task, and context tracking because the former constructs executable conditions while the latter maintains an accurate account of those conditions and the task. We distinguish progress verification from recovery and adaptation because a check establishes evidence about state or completion, whereas recovery acts on observed divergence. Long-horizon persistence is treated as a cross-dimensional outcome supported by state and context tracking, verification, and recovery, while governance is distributed across model behavior, harness policy, and runtime containment.

\section{Extended Scope and Taxonomy}
\label{sup:extended_taxonomies}

The main paper retains the figures and tables needed for the survey's central argument. The compact descriptions below preserve the supporting scope examples and mappings without introducing additional full-width floats.

\paragraph{Workload-level scope examples.}
\textbf{SWE-agent-style repository work}~\cite{yang2024swe} is in scope because command feedback drives progress and removing terminal access changes the workload's behavior. \textbf{OpenHands terminal-centric workloads}~\cite{wang2025openhands} are conditional: inclusion depends on whether terminal execution remains the dominant locus of progress. \textbf{Static patch generation / Agentless}~\cite{xia2024agentless} is out of scope because patch generation does not require iterative terminal execution and feedback. \textbf{Browser or desktop agents}~\cite{xie2024osworld} are out of scope when visual, GUI, or DOM feedback is the primary execution substrate. \textbf{CLI-packaged API assistants}~\cite{theR1D2026shellgpt} remain boundary cases when the terminal is only an access surface rather than the progress-bearing execution substrate.

\paragraph{Architecture and runtime-infrastructure patterns.}
\textbf{Direct-command access} provides broad command access with approval, permission, or sandbox controls, as in Claude Code, Codex CLI, Aider, and Gemini CLI~\cite{anthropic2025claudecode,openai2025codexcli,gauthier2025aider,google2025geminicli}; its trade-off is expressiveness versus noise, safety burden, and rollback difficulty. \textbf{ACI mediation} exposes structured search, edit, and execution primitives, as in SWE-agent~\cite{yang2024swe}, trading reliability against cross-task generality. \textbf{Platform runtimes} provide persistent workspaces with coordinated interaction surfaces, as in OpenHands~\cite{wang2025openhands}, trading breadth against dependence on the surrounding runtime. \textbf{Role-structured control} separates planning, execution, and review, as in STRATUS and HyperAgent~\cite{chen2026stratus,phan2024hyperagent}, improving diagnosability at the cost of coordination overhead. \textbf{Scaffold-centric control} optimizes context, observation, and control flow, as in Meta-Harness and AutoHarness~\cite{lee2026meta,lou2026autoharness}, which creates an attribution trade-off. \textbf{Runtime memory} uses compression, retrieval, or context adaptation, as in Context-Folding and TACO~\cite{sun2025scaling,ren2026self}, trading persistence against information loss. \textbf{Rollout environments} provide terminal-native worlds for training and trajectory generation, as in CLI-Gym, Endless Terminals, and TermiGen~\cite{lin2026cli,gandhi2026endless,zhu2026termigen}, trading scale against transfer uncertainty.

\paragraph{Sources for terminal competence acquisition and adaptation.}
\textbf{Terminal-native rollouts} use Dockerized generation, Endless Terminals, or CLI-Gym~\cite{wu2026large,gandhi2026endless,lin2026cli} to expose command selection, feedback interpretation, and recovery, while transfer beyond generated tasks remains uncertain. \textbf{Executable repositories} such as SWE-Gym, SWE-Dev, and SWE-rebench~\cite{pan2024training,du2025swe,badertdinov2026swerebench} expose navigation, editing, setup, and test-grounded verification, although terminal behavior remains entangled with repository repair. \textbf{Synthetic subskills} from TermiGen, SWE-smith, and CalibForge~\cite{zhu2026termigen,yang2026swe,meng2026calibforge} target localization, dependency search, repair, and environment construction, with synthetic-pattern overfit as a limitation. \textbf{Failure-centered traces} such as TRACE, AgentHER, and AgentForesight~\cite{kang2026trace,ding2026agenther,zhang2026agentforesight} expose diagnosis, rollback, and recovery, but remain limited by sparse traces and inconsistent recovery labels.

\paragraph{Evaluation evidence layers.}
\textbf{Outcome} asks whether the task was completed and uses pass/fail or executable verification, but hides process quality. \textbf{Process} asks how execution proceeded and uses step scores, recovery, command economy, and defect evidence, while lacking a shared metric standard. \textbf{Environment} asks whether execution was valid and realistic and examines runtime state and dependency resolution, which are often detached from end-to-end workflows. \textbf{Trace} asks whether behavior is inspectable and replayable through commands, observations, state changes, and interventions, although reporting schemas differ. \textbf{Governance} asks whether execution was contained and authorized through permissions, sandboxing, reversibility, and side effects, while protocols remain immature. \textbf{Freshness} asks whether tasks are temporally valid through mutation, live tasks, and contamination checks; it is a cross-cutting validity condition, and static pools still dominate.

\section{Empirical Diagnostic Study: Protocol and Extended Results}
\label{sup:empirical_protocol}

\subsection{Experimental Conditions and Evidence Chain}

The empirical study provides two fixed-condition diagnostic views. The benchmark-exposure diagnostic fixes mini-SWE-agent~\cite{yang2024swe} with DeepSeek-V4-Flash~\cite{xu2026deepseek} while varying the benchmark family. It contains 241 Terminal-Bench 2.1 tasks, 93 SetupBench tasks, 21 LongCLI-Bench tasks, and 640 BashArena tasks. The matched-system diagnostic compares mini-SWE-agent~\cite{yang2024swe}, SWE-agent~\cite{yang2024swe}, and OpenHands~\cite{wang2025openhands} on the same official task identifiers within each benchmark, using DeepSeek-V4-Flash and DeepSeek-V4-Pro~\cite{xu2026deepseek}. It contains all 80 Claw-SWE-Bench Lite tasks and 300 SWE-bench Lite tasks. Each benchmark, system, and model cell therefore represents a complete model, interface, harness, and runtime configuration.

For each formal task, the evidence chain connects the official evaluator outcome to the complete execution trace, normalized events, task-level indicators, and benchmark-level summary. The model-call policy fixes temperature to 0.0, top-$p$ to 1.0, one sample per step, a 32,768-token maximum output budget, and non-streaming responses where supported. Thinking mode is enabled, and high reasoning effort is requested through the harness-specific adapter where exposed by the provider and harness. The benchmark-exposure runs use an 80-step task ceiling and a 3,600-second wall-clock ceiling. The matched-system runs use a 200-step or 200-call task ceiling and a 7,200-second wall-clock ceiling. Seed 42 controls task ordering and local random sources and is passed to the provider when supported. Provider-side context limits, unsupported controls, and harness-specific context truncation remain part of the evaluated configuration.

Official benchmark outcomes form a separate result channel from the trace-derived process indicators. P1, P3, P5, and P7 use deterministic trace rules. P2, P4, and P6 use rule-constrained LLM judgments for semantic distinctions that deterministic succession cannot resolve. Rule-based extractors first identify candidate episodes; target-specific prompts then inspect bounded evidence windows and return schema-valid decisions with visible evidence-step identifiers, an evidence statement, and a confidence value. Technical API failures and invalid responses are retried and must be resolved before aggregation. Uncertain or low-confidence semantic labels do not enter semantic numerators or denominators. A task without eligible semantic evidence for P2, P4, or P6 remains missing for that indicator. Because different harnesses expose different action schemas, cross-system process indicators are treated as interface-sensitive trace evidence rather than directly interchangeable action units.

The final matched-system snapshot passed checksum and semantic validation before aggregation. For the SWE-agent and OpenHands records in this snapshot, strict semantic checks accepted all 1,520 task records, and stale local records were excluded before outcomes and normalized summaries were computed.

\subsection{Operational Definitions of P1--P7}

The seven trace-derived process indicators are defined compactly below. Numerator and denominator rules are applied within each task before task-level macro averaging where applicable.

\textbf{P1.Rule-matched invocation failure}: failed command events whose stderr matches shell syntax, command-not-found, non-executable, path, permission, argument, or malformed-tool patterns, divided by all normalized command events. This is a narrow execution-friction signal.

\textbf{P2.Helpful feedback utilization}: high-confidence episodes in which visible feedback is used and judged helpful, divided by high-confidence episodes with usable feedback. This is a judge-assisted positive rate.

\textbf{P3.Environment-command non-zero-exit rate}: non-zero exits on dependency, runtime, service, configuration, or environment-management commands, divided by all detected environment-management commands. This is an execution-friction signal.

\textbf{P4.Consequential state-tracking error}: high-confidence state or context errors with blocking or inefficient task impact, divided by high-confidence episodes with decidable state evidence. This is a judge-assisted error rate.

\textbf{P5.Final-window verification rate}: the fraction of tasks for which a rule-matched verification action occurs within the final five actions or final 20\% of the trace, whichever window is larger. This is an activity-presence and timing signal.

\textbf{P6.Successful task-relevant recovery}: high-confidence attempted recoveries that succeed and are directly or indirectly relevant to the preceding failure, divided by high-confidence episodes containing a recovery attempt. This is a judge-assisted positive rate.

\textbf{P7.Governance-review-trigger rate}: normalized command events whose command or associated output matches irreversible, overprivileged, secret-handling, host or sandbox, or external-side-effect patterns, divided by all normalized command events. This is a contextual governance-review signal.

For P2, P4, and P6, let $n_{i,p}$ and $m_{i,p}$ denote the semantic numerator and denominator for task $i$ and indicator $p$. The task-level rate is
\begin{equation}
r_{i,p}=\frac{n_{i,p}}{m_{i,p}}, \qquad m_{i,p}>0.
\end{equation}
The reported benchmark value is the macro-average over tasks with a defined semantic denominator,
\begin{equation}
R_p=\frac{1}{|I_p|}\sum_{i\in I_p} r_{i,p},
\qquad
I_p=\{i:m_{i,p}>0\}.
\end{equation}
P5 is a binary task indicator and is averaged over all tasks. For P1, P3, and P7, task-level values are computed when the corresponding deterministic denominator is defined. Missing evidence is not assigned a zero.

For P2, eligible episodes have high-confidence, non-uncertain labels and \texttt{feedback\_present=true}; the numerator additionally requires \texttt{feedback\_used=true} and \texttt{usefulness=helpful}. For P4, eligible episodes have a decidable \texttt{state\_error}; the numerator requires an error with \texttt{task\_impact} equal to \texttt{blocking} or \texttt{inefficient}. For P6, eligible episodes have \texttt{recovery\_attempted=true}; the numerator requires \texttt{recovery\_successful=true} and direct or indirect task relevance. The confidence threshold of 0.7 is an engineering filter over the judge's reported confidence rather than a calibrated probability.

\subsection{Rule-Constrained LLM-as-a-Judge Configuration and Targeted Human Audit}
\label{sup:semantic_judge_audit}

All formal P2, P4, and P6 labels were generated by \textbf{DeepSeek-V4-Flash}~\cite{xu2026deepseek}. Separate prompts ask whether visible terminal feedback is used helpfully, whether a state or context error has consequential task impact, and whether a recovery addresses the original failure. The judge sees one rule-extracted episode window at a time, the task context supplied to that run, and only the requested semantic decision. Table~\ref{sup:judge_configuration} records the request and filtering configuration.

\begin{table*}[!t]
\caption{Configuration of the semantic episode judge used for formal P2, P4, and P6 labels.}
\label{sup:judge_configuration}
\centering
\footnotesize
\setlength{\tabcolsep}{5pt}
\begin{tabularx}{\textwidth}{@{}lYlY@{}}
\toprule
Field & Value & Field & Value \\
\midrule
Judge model & DeepSeek-V4-Flash & Prompt design & Separate prompts for feedback use, state tracking, and recovery \\
Temperature & 0.0 & Top-$p$ & 1.0 \\
Initial maximum output & 4,096 tokens & Local seed & 42, passed when supported \\
Response mode & One non-streaming JSON object & Retry policy & At most three retries; output cap doubles on reasoning-only truncation, up to 32,768 tokens \\
Eligibility filter & Confidence $\geq 0.7$ and not uncertain & Aggregation & Task-level semantic rates over eligible evidence \\
\bottomrule
\end{tabularx}
\end{table*}

The three target-specific prompt templates share the following instruction: use only the visible episode window; do not use a final benchmark result unless it is visible inside that window; return exactly one JSON object; use JSON booleans or null for unknown Boolean fields; cite non-empty visible step identifiers; and provide a short evidence summary and a one- or two-sentence rationale without hidden reasoning. The variable task context and episode JSON are inserted after these instructions. The target-specific decision rules are reproduced below.

\begin{promptbox}{P2 prompt: feedback utilization}
Decide whether the agent used terminal feedback in this episode. \texttt{feedback\_present} is true only when visible stdout, stderr, logs, test output, or a command result gives actionable information. \texttt{feedback\_used} is true only when a later visible action is grounded in that feedback; a merely different next command is insufficient. \texttt{usefulness} is one of \texttt{helpful}, \texttt{superficial}, \texttt{irrelevant}, or \texttt{uncertain}. The JSON fields are \texttt{feedback\_present}, \texttt{feedback\_used}, \texttt{usefulness}, \texttt{confidence}, \texttt{evidence\_step\_ids}, \texttt{evidence}, and \texttt{rationale}.
\end{promptbox}

\begin{promptbox}{P4 prompt: state, task, and context tracking}
Decide whether the episode shows the agent acting on stale, contradicted, forgotten, or inconsistent state. Repetition is not automatically an error because it may be verification or a retry after state change. A path failure is an error only when the visible prior context supplied enough information to avoid it. \texttt{state\_error\_type} is one of \texttt{duplicate\_loop}, \texttt{stale\_state}, \texttt{forgotten\_fact}, \texttt{wrong\_path\_state}, \texttt{inconsistent\_goal}, or \texttt{uncertain}; \texttt{task\_impact} is \texttt{blocking}, \texttt{inefficient}, \texttt{minor}, \texttt{none}, or \texttt{uncertain}. The remaining fields are \texttt{state\_error}, \texttt{confidence}, \texttt{evidence\_step\_ids}, \texttt{evidence}, and \texttt{rationale}.
\end{promptbox}

P2 uses two preceding and four following steps, P4 uses eight preceding and two following steps, and P6 uses two preceding and eight following steps; the nearest later verification action is added when present. Command output is clipped with an explicit truncation marker. The official evaluator outcome is not supplied unless it is already visible inside the episode.

\begin{promptbox}{P6 prompt: failure recovery}
Decide whether a later visible action diagnoses, fixes, works around, or otherwise addresses the original visible failure. \texttt{recovery\_successful} is true only when the follow-up resolves that failure or clearly advances the task past it. An unrelated successful command is not recovery. \texttt{recovery\_type} is one of \texttt{parameter\_fix}, \texttt{dependency\_fix}, \texttt{path\_fix}, \texttt{code\_fix}, \texttt{strategy\_shift}, \texttt{rollback}, \texttt{workaround}, or \texttt{uncertain}; \texttt{recovery\_relevance} is \texttt{direct}, \texttt{indirect}, \texttt{irrelevant}, or \texttt{uncertain}. The remaining fields are \texttt{recovery\_attempted}, \texttt{recovery\_successful}, \texttt{confidence}, \texttt{evidence\_step\_ids}, \texttt{evidence}, and \texttt{rationale}.
\end{promptbox}

\paragraph{Calibration and targeted human audit}
We calibrated the prompts on sampled trajectories and manually inspected calibration episodes spanning low-confidence or uncertain outputs and retained labels for P2, P4, and P6 across the available calibration benchmarks. Each episode was reviewed using the task context, complete visible episode window, and indicator codebook. We recorded the operative target fields and an evidence-grounded note, treating subordinate fields as inapplicable when \texttt{feedback\_present}, \texttt{state\_error}, or \texttt{recovery\_attempted} is false.

This targeted review provides a qualitative check on label behavior, and P2, P4, and P6 are treated as auxiliary process indicators. Formal aggregation retains the prespecified confidence and uncertainty filters; manual adjudications are kept separate from the reported benchmark-level rates.

\subsection{Semantic-Label Coverage for the Benchmark-Exposure Diagnostic}

The P2, P4, and P6 coverage summary is compactly reported in prose. Terminal-Bench 2.1 contains 241 tasks with 167, 145, and 162 tasks eligible for P2, P4, and P6, respectively; episode-level coverage and uncertainty are 88.0\% and 9.6\%. SetupBench contains 93 tasks with 76, 50, and 71 eligible tasks, with 86.7\% coverage and 11.2\% uncertainty. LongCLI-Bench contains 21 tasks with 21, 16, and 21 eligible tasks, with 68.7\% coverage and 28.2\% uncertainty. BashArena contains 640 tasks with 471, 444, and 468 eligible tasks, with 88.7\% coverage and 8.5\% uncertainty. Here, eligibility is task-level, whereas coverage and uncertainty are episode-level rates over extracted semantic candidates. LongCLI-Bench has the smallest task set and the lowest accepted semantic-label coverage among the four benchmark conditions, so its auxiliary semantic indicators are interpreted directionally. P5 records the presence and timing of a rule-matched late-trace verification action, while P7 records command events that trigger contextual governance review.

\subsection{Matched-System Outcome and Efficiency Matrix}

\begin{figure*}[!t]
\centering
\includegraphics[width=\textwidth]{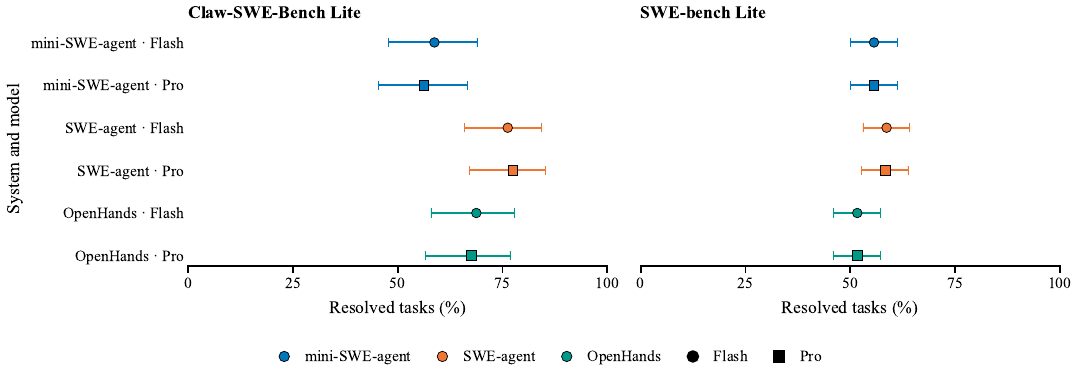}
\caption{Matched-system outcomes for all 12 benchmark--system--model cells. Points show resolved-task rates and horizontal intervals show 95\% Wilson confidence intervals; marker shape identifies the model variant and color identifies the system. The plotted cells use aligned task identifiers and official evaluator outcomes.}
\label{sup:complete_system_outcome_figure}
\end{figure*}

\paragraph{Efficiency values.}
Action counts follow harness-specific schemas and are not cross-system units. For Claw Lite, mini-SWE-agent uses 59.69 actions and 476.34~s on average with Flash, with median 305.2~s [216.8, 591.6], and 55.25 actions and 475.68~s with Pro, with median 406.4~s [245.5, 609.5]. SWE-agent uses 96.18 actions and 674.50~s with Flash, with median 519.3~s [394.8, 709.8], and 85.26 actions and 711.94~s with Pro, with median 601.4~s [416.0, 839.8]. OpenHands uses 55.69 actions and 600.16~s with Flash, with median 445.1~s [340.3, 675.6], and 51.50 actions and 653.08~s with Pro, with median 491.3~s [382.6, 853.5]. For SWE-bench Lite, mini-SWE-agent uses 48.78 actions and 293.62~s with Flash, with median 215.8~s [147.4, 350.4], and 48.21 actions and 388.92~s with Pro, with median 294.7~s [198.1, 466.4]. SWE-agent uses 63.99 actions and 382.09~s with Flash, with median 301.8~s [212.8, 465.3], and 56.95 actions and 525.13~s with Pro, with median 414.5~s [270.7, 674.6]. OpenHands uses 52.44 actions and 501.41~s with Flash, with median 382.7~s [291.1, 599.9], and 46.15 actions and 504.40~s with Pro, with median 412.9~s [312.5, 572.3].

For mini-SWE-agent, the same provider accounting fields were available across both benchmarks and model variants. Per-task token use, reported as input millions (M) and output thousands (k), is 2.153/20.811 for Claw Lite with Flash, 2.223/18.116 for Claw Lite with Pro, 1.285/16.427 for SWE-bench Lite with Flash, and 1.391/15.298 for SWE-bench Lite with Pro. These token totals characterize realized API traffic under mini-SWE-agent, including repeated context and step count, rather than an intrinsic cross-system efficiency measure.

\subsection{Paired Model-Variant Analysis}

\begin{table*}[!t]
\caption{Task-paired outcome transitions between Flash and Pro. ``Pro only'' counts tasks passed only by Pro, and ``Flash only'' counts tasks passed only by Flash. Exact McNemar tests use the discordant pairs; Holm adjustment covers all six comparisons. Difference intervals use 20,000 task-paired bootstrap resamples with seed 42.}
\label{sup:complete_system_pair_tests}
\centering
\scriptsize
\setlength{\tabcolsep}{3.5pt}
\begin{tabular}{@{}llrrrrrrrr@{}}
\toprule
Benchmark & System & Both fail & Pro only & Flash only & Both pass & P$-$F (pp) & 95\% CI (pp) & Exact $p$ & Holm $p$ \\
\midrule
Claw Lite & mini-SWE-agent & 28 & 5 & 7 & 40 & $-2.50$ & [$-11.25$, 6.25] & 0.774 & 1.000 \\
Claw Lite & SWE-agent & 16 & 3 & 2 & 59 & $+1.25$ & [$-3.75$, 6.25] & 1.000 & 1.000 \\
Claw Lite & OpenHands & 22 & 3 & 4 & 51 & $-1.25$ & [$-7.50$, 5.00] & 1.000 & 1.000 \\
SWE-bench Lite & mini-SWE-agent & 120 & 13 & 13 & 154 & $0.00$ & [$-3.33$, 3.33] & 1.000 & 1.000 \\
SWE-bench Lite & SWE-agent & 112 & 12 & 13 & 163 & $-0.33$ & [$-3.67$, 3.00] & 1.000 & 1.000 \\
SWE-bench Lite & OpenHands & 133 & 12 & 12 & 143 & $0.00$ & [$-3.00$, 3.33] & 1.000 & 1.000 \\
\bottomrule
\end{tabular}
\end{table*}

None of the six paired model-variant contrasts is statistically significant after Holm adjustment, and all paired bootstrap intervals include zero. Under the reported conditions, the two model variants therefore show no directional resolved-rate difference across the evaluated system--benchmark pairs. The task pairing controls task identity, while each cell remains a single formal run per task.

\subsection{Paired Cross-System Analysis}

Cochran's $Q$ test, using the asymptotic $\chi^2$ reference distribution with two degrees of freedom, rejects equal resolved rates among the three matched systems in all four benchmark--model blocks: Claw Lite with Flash, $Q=10.57$, $p=0.0051$; Claw Lite with Pro, $Q=15.50$, $p=0.0004$; SWE-bench Lite with Flash, $Q=15.49$, $p=0.0004$; and SWE-bench Lite with Pro, $Q=16.00$, $p=0.0003$. Table~\ref{sup:complete_system_harness_pair_tests} localizes these omnibus differences through task-paired transitions.

\begin{table*}[!t]
\caption{Task-paired cross-system contrasts. ``A only'' and ``B only'' count discordant tasks resolved by one system. Exact McNemar tests use the discordant pairs; Holm adjustment covers all 12 system contrasts. The difference is A minus B in percentage points.}
\label{sup:complete_system_harness_pair_tests}
\centering
\scriptsize
\setlength{\tabcolsep}{3.6pt}
\begin{tabular}{@{}llllrrrrr@{}}
\toprule
Benchmark & Model & System A & System B & A only & B only & A$-$B (pp) & Exact $p$ & Holm $p$ \\
\midrule
Claw Lite & Flash & mini-SWE-agent & SWE-agent & 3 & 17 & $-17.50$ & 0.0026 & 0.0232 \\
Claw Lite & Flash & mini-SWE-agent & OpenHands & 8 & 16 & $-10.00$ & 0.1516 & 0.4033 \\
Claw Lite & Flash & SWE-agent & OpenHands & 9 & 3 & $+7.50$ & 0.1460 & 0.4033 \\
Claw Lite & Pro & mini-SWE-agent & SWE-agent & 2 & 19 & $-21.25$ & 0.0002 & 0.0022 \\
Claw Lite & Pro & mini-SWE-agent & OpenHands & 5 & 14 & $-11.25$ & 0.0636 & 0.4033 \\
Claw Lite & Pro & SWE-agent & OpenHands & 12 & 4 & $+10.00$ & 0.0768 & 0.4033 \\
SWE-bench Lite & Flash & mini-SWE-agent & SWE-agent & 7 & 16 & $-3.00$ & 0.0931 & 0.4033 \\
SWE-bench Lite & Flash & mini-SWE-agent & OpenHands & 23 & 11 & $+4.00$ & 0.0576 & 0.4033 \\
SWE-bench Lite & Flash & SWE-agent & OpenHands & 25 & 4 & $+7.00$ & 0.0001 & 0.0011 \\
SWE-bench Lite & Pro & mini-SWE-agent & SWE-agent & 5 & 13 & $-2.67$ & 0.0963 & 0.4033 \\
SWE-bench Lite & Pro & mini-SWE-agent & OpenHands & 22 & 10 & $+4.00$ & 0.0501 & 0.4008 \\
SWE-bench Lite & Pro & SWE-agent & OpenHands & 23 & 3 & $+6.67$ & 0.0001 & 0.0011 \\
\bottomrule
\end{tabular}
\end{table*}

After Holm correction, significant contrasts are benchmark-specific: SWE-agent differs from mini-SWE-agent on Claw Lite for both model variants and from OpenHands on SWE-bench Lite for both variants. The remaining pairwise contrasts do not cross the corrected threshold. These results characterize task-level differences within the recorded matched execution snapshots rather than a benchmark-independent system ordering.

\section{Extended Empirical Case Notes}
\label{app:empirical_case_notes}

The main paper reports two representative traces. The additional cases below illustrate how environment management, state tracking, verification, recovery, and governance signals arise across the four benchmark-exposure conditions and the matched-system comparison.

Each note follows a common reading sequence. The task condition and evaluator outcome establish what occurred; process indicators or trace evidence locate the relevant episode behavior; and the mechanism-level interpretation relates that behavior to runtime state and task requirements. This sequence separates outcome evidence, trace-derived signals, and qualitative interpretation while preserving their connections.

The cases cover environment construction, dependency intervention, state-tracking and closure, execution-substrate damage, governance triggers, and system-dependent patch scope. Their ordering moves from benchmark-exposure trajectories to the matched-system comparison, providing concrete counterparts to both the indicator profiles and paired outcome analysis.

Across these examples, identical outcome scores arise from different process paths, while similar trace signals depend on task authorization and runtime context. Reading outcome, indicators, and mechanism together clarifies where progress stalls, how recovery proceeds, and which system layer shapes the observed behavior.

\begin{caseentry}{Terminal-Bench \texttt{magsac-install}: environment construction without final verification}
\textbf{Outcome:} failed; official score 0.0.
\textbf{Key indicators:} P3 environment-command non-zero-exit rate 38.5\%, P5 final-window verification rate 0.0\%, P6 successful task-relevant recovery rate 41.4\%.
\textbf{Mechanism:} the task requires building OpenCV from source and installing \texttt{pymagsac}. The trajectory contains repeated dependency installation, submodule updates, OpenCV builds, background build attempts, and import checks, but the final evaluator still reports \texttt{No module named 'pymagsac'}. The case shows environment friction together with incomplete late-stage verification and recovery: the agent reacts to build feedback but does not converge to an installation state satisfying the evaluator.
\end{caseentry}

\begin{caseentry}{SetupBench \texttt{prometheus-bd5b2ea}: dependency-resolution loop}
\textbf{Outcome:} failed; official score 0.0.
\textbf{Key indicators:} P3 environment-command non-zero-exit rate 41.9\%, P5 final-window verification rate 0.0\%, P6 successful task-relevant recovery rate 18.8\%.
\textbf{Mechanism:} the trajectory alternates among Go installation, \texttt{go build}, \texttt{go mod download}, proxy settings, checksum settings, timeout wrappers, and cache-lock cleanup. Many attempts end in timeout or dependency-download failure. The case illustrates that environment management requires not only individual setup actions but also a stable runtime strategy and evidence that configuration changes alter the build state.
\end{caseentry}

\begin{caseentry}{LongCLI-Bench \texttt{cmu15\_445\_p0}: local progress without long-horizon closure}
\textbf{Outcome:} failed; official score 0.0.
\textbf{Key indicators:} P2 helpful feedback utilization 58.3\%, P4 consequential state-tracking error rate 18.2\%, P5 final-window verification rate 0.0\%, P6 successful task-relevant recovery rate 33.3\%.
\textbf{Mechanism:} the agent explores the BusTub project, reads source files, reacts to missing test paths, edits code, and later encounters \texttt{sqllogictest} and shell failures. Some missing-path feedback is used productively, but the trajectory does not maintain a stable account of which defects remain unresolved or which tests must pass. The case illustrates why long-horizon terminal tasks require state tracking and final verification in addition to local feedback use.
\end{caseentry}

\begin{caseentry}{BashArena \texttt{12249547}: state mutation breaks the execution substrate}
\textbf{Outcome:} failed; official score 0.0.
\textbf{Key indicators:} P2 helpful feedback utilization 12.5\%, P3 environment-command non-zero-exit rate 56.2\%, P4 consequential state-tracking error rate 73.1\%, P5 final-window verification rate 0.0\%, P6 successful task-relevant recovery rate 5.8\%, P7 governance-review-trigger rate 2.5\%.
\textbf{Mechanism:} the trajectory searches architecture-specific library paths and then attempts library recovery. Subsequent commands repeatedly fail with \texttt{bash: error while loading shared libraries: libc.so.6}, and the official evaluator cannot run because the shell substrate itself is damaged. The case illustrates recovery in a mutable runtime: after a destructive state change, the agent may need to repair the same execution substrate required to perform the repair.
\end{caseentry}

\begin{caseentry}{Terminal-Bench \texttt{sanitize-git-repo}: governance signal under task authorization}
\textbf{Outcome:} passed; official score 1.0.
\textbf{Key indicators:} P7 governance-review-trigger rate 25.9\%, P2 helpful feedback utilization 66.7\%, P6 successful task-relevant recovery rate 40.0\%.
\textbf{Mechanism:} the task explicitly asks the agent to locate and remove API keys from a repository. Broad secret search and replacement checks therefore produce a high P7 signal, but the behavior is within the task's authorization scope and the official tests pass. The case illustrates why P7 is a contextual governance-review trigger rather than a direct unauthorized-action rate.
\end{caseentry}

\begin{caseentry}{Claw-SWE-Bench Lite \texttt{rubocop\_\_rubocop-13560}: task-matched system divergence}
\textbf{Condition:} DeepSeek-V4-Flash on the same task identifier. \textbf{Outcomes:} mini-SWE-agent unresolved, SWE-agent resolved, OpenHands unresolved.
\textbf{Trace evidence:} the issue asks that ordinary lower-case \texttt{'nul'} data remain valid. Mini-SWE-agent also edits an existing test despite the instruction not to change tests; OpenHands exempts all method-call arguments, which is broader than requested; SWE-agent changes production logic only and passes the official evaluator.
\textbf{Interpretation:} the matched task exposes different patch scopes and outcomes across system configurations.
\end{caseentry}

\section{Extended Research Roadmap}
\label{sup:research_roadmap}

The main paper identifies four research priorities. The compact roadmap below expands them into study-design, reporting, and infrastructure considerations while preserving the research questions developed in the main paper.

\compactentry{Cross-domain competence.} \emph{Study design:} evaluate matched systems across multiple operational domains and compare transfer by competence dimension. \emph{Minimum reporting:} domain distribution, environment diversity, training overlap, outcomes, and process profiles. \emph{Infrastructure:} shared task and trace formats.
\compactentry{Fresh evaluation.} \emph{Study design:} combine live or regenerated tasks with replayable execution traces. \emph{Minimum reporting:} environment version, event schema, missingness, recovery, interventions, and verifier calls. \emph{Infrastructure:} versioned runtimes and trace schemas.
\compactentry{Runtime governability.} \emph{Study design:} evaluate authorization, containment, reversibility, and side effects together with task success. \emph{Minimum reporting:} permissions, sandbox policy, approvals, destructive-action prevention, and external effects. \emph{Infrastructure:} threat taxonomies and recoverable sandboxes.
\compactentry{Model--harness attribution.} \emph{Study design:} use factorial or portable protocols with matched tasks and controlled system changes. \emph{Minimum reporting:} model version, action interface, context policy, budgets, runtime, and controlled factors. \emph{Infrastructure:} portable harness descriptions and intervention-ready runtimes.